\documentclass[11pt]{article}
\usepackage[linesnumbered,ruled,vlined]{algorithm2e}
\usepackage{amsfonts}
\usepackage{amsmath}
\usepackage{amssymb}
\usepackage{authblk}
\usepackage{bm}
\usepackage{booktabs}
\usepackage[margin=1in]{geometry}
\usepackage{graphicx}
\usepackage[separate-uncertainty=true]{siunitx}
\usepackage[colorlinks=true]{hyperref}
\usepackage{appendix}
\usepackage[]{mdframed}
\usepackage{float}
\usepackage{lineno}
\usepackage{multirow}

\title{FMEnets:  Flow, Material, and  Energy networks for non-ideal plug flow reactor design}
\author[1]{Chenxi Wu}
\author[2]{Juan Diego Toscano}
\author[2]{Khemraj Shukla}
\author[3]{Yingjie Chen}
\author[3]{Ali Shahmohammadi}
\author[3]{Edward Raymond}
\author[3]{Thomas Toupy}
\author[3]{Neda Nazemifard}
\author[3]{Charles Papageorgiou}
\author[2,*]{George Em Karniadakis}
\affil[1]{School of Engineering, Brown University, Providence, RI 02912, USA}
\affil[2]{Division of Applied Mathematics, Brown University, Providence, RI 02912, USA}
\affil[3]{Process Engineering and Technology, Synthetic Molecule Process Development, Takeda Pharmaceuticals, Cambridge, MA 02142, USA 
}
\affil[*]{Corresponding author. Email: george\_karniadakis@brown.edu}
\date{}

\begin{document}

\maketitle

\begin{abstract}
We propose FMEnets, a physics-informed machine learning framework for the design and analysis of non-ideal plug flow reactors. FMEnets integrates the fundamental governing equations (Navier–Stokes for fluid flow, material balance for reactive species transport, and energy balance for temperature distribution) into a unified multi-scale network model. The framework is composed of three interconnected sub-networks with independent optimizers that enable both forward and inverse problem-solving. In the forward mode, FMEnets predicts velocity, pressure, species concentrations, and temperature profiles using only inlet and outlet information. In the inverse mode, FMEnets utilizes sparse multi-residence-time measurements to simultaneously infer unknown kinetic parameters and states. FMEnets can be implemented either as FME-PINNs, which employ conventional multilayer perceptrons, or as FME-KANs, based on Kolmogorov–Arnold Networks. Comprehensive ablation studies highlight the critical role of the FMEnets architecture in achieving accurate predictions. Specifically, FME-KANs are more robust to noise than FME-PINNs, although both representations are comparable in accuracy and speed in noise-free conditions. The proposed framework is applied to three different sets of reaction scenarios and is compared with finite element simulations. FMEnets effectively captures the complex interactions, achieving relative errors less than 2.5\% for the unknown kinetic parameters. The new network framework not only provides a computationally efficient alternative for reactor design and optimization, but also opens new avenues for integrating empirical correlations, limited and noisy experimental data, and fundamental physical equations to guide reactor design.
\end{abstract}


\section{Introduction}
The chemical reactor plays a key role in most chemical processes \cite{fuad2015systematic}, as it enables the production of valuable products that are otherwise unavailable in nature, such as polymers and medications \cite{sheldon2012fundamentals}. Specificaly, reactor design is essential in any industry involving chemical reactions \cite{froment1979chemical}, including the polymer industry, dye and pigment manufacturing, the color industry, and the pharmaceutical industry. Reactor  design and operation are essential in pharmaceutical manufacturing for synthesizing chemical compounds, growing microorganisms, and separating mixtures. Effective reactor design typically results in substantial cost savings and revenue opportunities for chemical plants \cite{towler2021chemical}. To satisfy the growing demand for pharmaceuticals in both healthcare and everyday life, chemical engineers strive to develop optimal strategies for scaling up various reactor types to produce a wide range of pharmaceutical products efficiently. Furthermore, reactor design is crucial for operational safety. The appropriate reactor design can reduce accident risks, ensure controlled reactions, and incorporate robust safety measures to protect personnel and the surrounding environment \cite{stoessel2021thermal}.

The complexity of a reactor design problem can vary significantly based on the assumptions made and the degree of deviation from ideal conditions. These deviations can arise from non-ideal flow conditions, non-isothermal operations, complex reaction kinetics, mass transfer limitations in heterogeneous reactions, and non-negligible pressure drop \cite{fogler2010essentials}. Traditionally, the underlying mechanism of these complex multiphase characteristics can be studied and understood by experimental and numerical techniques. In particular, computational fluid dynamics (CFD) is an effective numerical tool to describe detailed flow and transport behavior \cite{fogler2010essentials}. Historically, data from these experimental and numerical techniques have been used to calibrate or develop simplified engineering correlations (e.g., the Ergun equation and Gunn heat transfer correlations) that are useful for engineering design, optimization, and control \cite{zhu2022review}. However, conventional reactor design methods have limitations that limit their adaptability and accuracy. In particular, the use of simplifying assumptions can compromise the physical realism of models, while the choice of models often depends on the judgment and experience of the engineers \cite{fogler2010essentials}. In addition, these methods frequently depend on empirical relationships drawn from limited experimental data, which constrains their applicability to different reaction systems and scaled-up conditions. Numerical approaches, such as finite element or finite difference methods, can be computationally expensive, limiting the scope of rapid iteration and design optimization \cite{zhu2022review}. Consequently, traditional reactor design strategies may struggle to keep up with the evolving complexities of modern chemical processes.

Machine learning (ML) methods offer a promising alternative to traditional reactor design techniques by reducing computational cost and integrating diverse data sources \cite{haghighatlari2019advances,wu2023application}. Depending on the amount of information or domain knowledge available, ML models can be categorized into three primary types: pure data-driven methods (black-box models), hybrid methods (grey-box models) that utilize ML to partially replace or optimize traditional physical model \cite{ghasem2024combining,li2023analysis}, and domain knowledge-informed ML methods that integrate prior knowledge of physics, mathematical laws, chemical mechanisms, or boundary conditions as constraints within the ML algorithm's architecture \cite{zhu2022review,cao2024laplace,wang2021learning}.  Physics-informed neural networks (PINNs) \cite{karniadakis2021physics,raissi2019physics,lu2021deepxde} is an approach in this last category, as PINNs can incorporate and integrate physical laws and experimental data seamlesssly. PINNs have been widely used in various computational science and engineering problems \cite{cuomo2022scientific,toscano2025pinns,hao2022physics}, including in materials \cite{chen2020physics,niu2023modeling}, fluid dynamics \cite{raissi2020hidden,cai2021physics,zhang2023artificial,sharma2023review}, biology \cite{yazdani2020systems,daneker2023systems,zapf2022investigating}, topology optimization \cite{lu2021physics}, system engineering \cite{wang2024digital,yang2024data}, geophysics \cite{ding2023self}, and reactor design \cite{patel2023optimal,cohen2024data,choi2022physics,li2024unit}. 

In chemical engineering, extensive research has focused on applying physics-informed machine learning methods to tackle the challenges of multiphase flow \cite{qiu2022physics}, chemical reactions \cite{lastrucci2024physics,wu2024physics,hou2025physics,sun2023physics}, and heat transfer \cite{peng2023physics,laubscher2021simulation}. Despite the significant advancements of machine learning in numerous fields \cite{rajendra2022advancement, brunton2020machine, wei2019machine, sidey2019machine, wu2023gpt}, its implementation in reactor design still faces challenges from two aspects: data and model development \cite{zhu2022review}. On the data side, limited datasets can reduce the effectiveness of purely data-driven methods, making the integration of prior knowledge essential. Data gathered from different experiments or simulations can be inconsistent or vary in accuracy. Thus, a machine learning model in reactor design must be robust while handling noise and uncertainty. Moreover, model selection involves maintaining a balance between computational cost and predictive accuracy. Furthermore, inferring unknown reaction parameters when the reaction kinetics are not fully established is critical to ensure reliable predictions even with limited data.

This work presents an integrated neural network framework, FMEnets, which stands for flow, material, and energy balance networks. FMEnets architecture comprises three sub-networks with independent optimizers, and can be applied in both forward and inverse modes based on the sequential training schedule. In the forward mode,  FMEnets rely only on inlet and outlet information to predict velocities, pressure, molar concentrations, and temperature profiles along the reactor. In the inverse mode, FMEnets incorporate sparse experimental observations from reactors operating at different residence times, hence enabling the simultaneous inference of unknown kinetic parameters in addition to the relevant flow and transport variables. Two principal implementations of FMEnets are introduced: FME-PINNs, which employ conventional multilayer perceptrons (MLPs), and FME-KANs, which utilize Kolmogorov-Arnold Networks (KANs). 

\subsection{Related Work and Our Contributions}
In recent years, several studies have used physics-informed neural networks (PINNs) to model and optimize plug-flow reactors. Ngo et al. \cite{ngo2022forward} developed forward PINNs using one-dimensional ($z$) design equations for plug flow reactors, where $z$ is the axial position. Sun et al. \cite{sun2023physics} proposed PINNs for two-dimensional ($t$, $x$) material balance equations, solving both forward and inverse problems in nonlinear reacting flows. Here, $t$ represents time, and $x$ is the spatial domain. Patel et al. \cite{patel2023optimal} employed PINNs to analyze a tubular reactor with two-dimensional ($r$, $z$) material balances, where $r$ is the radial position and $z$ is the axial position. The study also optimizes temperature trajectories to maximize reactor yields. Cohen et al. \cite{cohen2024data} apply PINNs to a plug flow reactor using two-dimensional ($t$, $x$) material balance equations, comparing the performance of PINNs and symbolic regression in identifying unknown parameters.

Except for Patel et al. \cite{patel2023optimal}, these past works have considered one spatial dimension. Although Patel et al. \cite{patel2023optimal} addressed a two-dimensional setup, this study assumes an empirical temperature profile instead of solving an energy balance. Consequently, these investigations often employed simplified reactor models that may not fully capture the complexities of real-world conditions, such as non-ideal temperature distributions and other intricate factors. Accurately representing these phenomena is essential for ensuring safe operation and achieving efficient optimization.

To address these challenges, we introduce FMEnets, a neural network framework that integrates the Navier–Stokes equations, material balances, and an energy balance to solve non-ideal plug flow reactor design problems in forward and inverse modes. FMEnets does not assume that reactor behavior follows the design equations for the ideal reactor; instead, it allows experimental data to guide the underlying physics, thereby accommodating non-idealities. This framework offers a realistic, high-dimensional model without preset temperature or velocity profiles. Furthermore, by incorporating Kolmogorov-Arnold Networks, FMEnets demonstrates robust performance in the presence of noisy experimental data. Additionally, the model can infer unknown kinetic parameters when supplementary experimental data are available, making it well-suited for applications where kinetic information is incomplete.

\section{Problem Setup}
\subsection{Steady-State Non-adiabatic Plug Flow Reactors with Coolant}
\label{Plug_flow_react}

\begin{figure}[htbp]
    \centering
    \includegraphics[scale = 0.6]{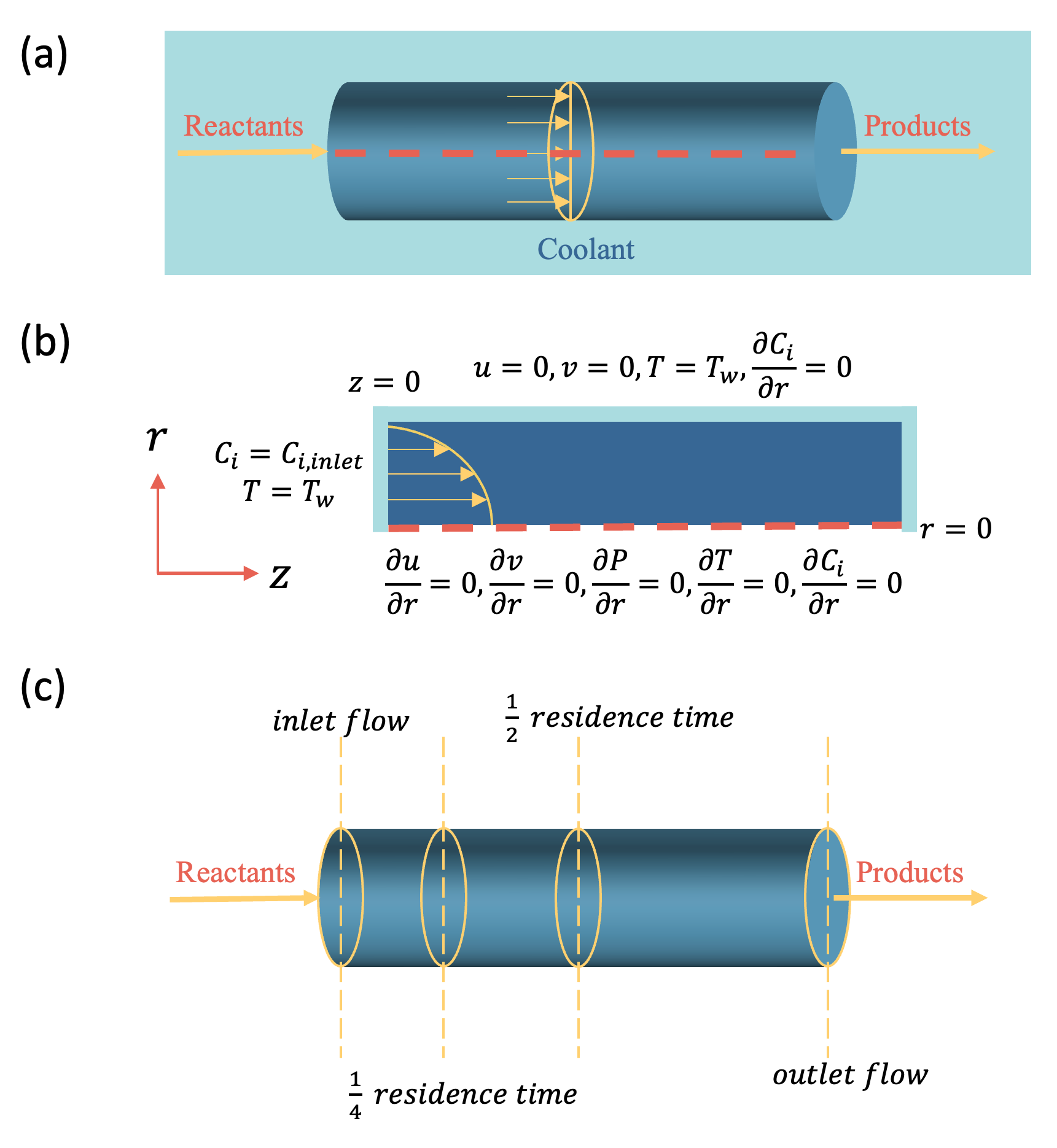}
    \caption{(a) Schematic of the tubular reactor with external coolant; (b) Modeled domain and boundary conditions; and (c) Sampling/measurement points at different residence times along the reactor.}
    \label{fig:pfr}
\end{figure}

In this study, we investigate steady-state, non-adiabatic plug-flow reactors (PFRs) carrying out exothermic reactions in the presence of a coolant, as shown in Figure \ref{fig:pfr}a. These continuous-flow reactors are immersed in an isothermal bath, enabling heat removal to maintain the reactor wall at a constant temperature of 400K, which is identical to the coolant temperature. However, the temperature within the reactor is not uniform. This design offers improved temperature control, increased safety, and enhanced selectivity and yield, especially in highly exothermic reactions. Consequently, it is widely used in various pharmaceutical processes.

To illustrate the performance and versatility of this reactor configuration, we examine three different reaction systems in steady-state, non-adiabatic PFRs with coolant:

\paragraph{Case 1: Two-Component Reaction System.} \label{para: case1}
For Case 1, \( i \in \{A, B\} \), and the reaction pathway is given by
\[
A \to B.
\]
The net rates of consumption for each species are
\[
r_A = k_1 C_A^2, \quad r_B = -k_1 C_A^2,
\]
where \( k_1 \) is the rate constant.

\paragraph{Case 2: Three-Component Reaction System.} \label{para: case2}
For Case 2, \( i \in \{A, B, C\} \), two sequential reactions,
\[
\text{A} \rightarrow \text{B},
\]
followed by
\[
\text{B} \rightarrow \text{C},
\]
\noindent are modeled following the approach of \cite{patel2023optimal}. Here, species A is the only feed, with B and C forming as the reaction proceeds downstream.

The net rates of consumption for each species are
\[
r_A = k_1 C_A^2, \quad 
r_B = -k_1 C_A^2 + k_2 C_B^2, \quad 
r_C = -k_2 C_B^2,
\]
where \( k_1 \) and \( k_2 \) are the respective rate constants for the two steps.

\paragraph{Case 3: Six-Component Reaction System.} \label{para: case3}
For Case 3, \( i \in \{A, B, C, D, E, F\} \), and three simultaneous reactions are considered:
\[
B + C \to D, \quad A + B \to E, \quad A + B \to F.
\]
The net rates of consumption for these species are given by:
\[
\begin{aligned}
r_A &= k_2 C_B C_C + k_3 C_A C_B, \\
r_B &= k_1 C_B C_C + k_2 C_A C_B + k_3 C_A C_B, \\
r_C &= k_1 C_B C_C, \\
r_D &= - k_1 C_B C_C, \\
r_E &= - k_2 C_A C_B, \\
r_F &= - k_3 C_A C_B,
\end{aligned}
\]
\noindent where \( k_1 \), \( k_2 \), and \( k_3 \) represent the rate constants for each of the parallel reactions. In this case, species A, B, and C are fed into the reactor, yielding D, E, and F along the reactor’s length.

\subsection{Non-dimensionalized Governing Equations}
\label{Governing_Eqs}
In this study, we consider the reactor to be axisymmetric, with negligible radial and axial velocities. We assume fully developed flow in the liquid phase, which is modeled as Newtonian fluid. The system operates under steady-state conditions. Furthermore, the reactants and products are present at sufficiently low concentrations such that the solvent properties dominate; hence, the solvent's heat capacity, density, viscosity, and thermal conductivity are treated as constants throughout. 

The reactor design problem often spans multiple scales, requiring a robust modeling approach. Scaling and non-dimensionalization enhance the numerical stability and convergence speed of PINNs. Therefore, we solve the equations in a dimensionless form involving three dimensionless numbers: the Reynolds number ($Re$),
\[
Re = \frac{\rho V D}{\mu},
\]
the mass-transfer Péclet number ($Pe$),
\[
Pe = \frac{V D}{D_{AB}},
\]
and the thermal Péclet number ($Pe_T$),
\[
Pe_T = \frac{V D}{\alpha}.
\]
Here, \(\rho\) is the fluid density, \(V\) is the characteristic velocity, \(D\) is a characteristic length (e.g., the diameter of the plug flow reactor), \(\mu\) is the dynamic viscosity, \(D_{AB}\) is the diffusion coefficient, and the thermal diffusivity ($\alpha$),
\[
\alpha = \frac{k_c}{\rho C_p},
\]
where \(k_c\) is the thermal conductivity and \(C_p\) the specific heat capacity.

With the assumptions above, we simplified the governing equations and rewrote them in non-dimensionalized residual forms as follows.

\subsubsection{Navier-Stokes Equations}
The plug flow reactor is typically modeled in cylindrical coordinates. We note that \( u \) represents the velocity in the \( z \)-direction, while \( v \) represents the velocity in the \( r \)-direction. In dimensionless form, the Navier–Stokes equations can be written as:
\begin{align}
\label{NS1}
e_{NS1} &=  u\frac{\partial u}{\partial z} +  v\frac{\partial u}{\partial r} - \frac{1}{Re} \left( \frac{1}{r} \frac{\partial u}{\partial r} + \frac{\partial^2 u}{\partial r^2}  + \frac{\partial^2 u}{\partial z^2} \right) + \frac{\partial P}{\partial z} = 0 \\
\label{NS2}
e_{NS2} &=  u\frac{\partial v}{\partial z} + v\frac{\partial v}{\partial r} - \frac{1}{Re} \left( \frac{1}{r} \frac{\partial v}{\partial r} +  \frac{\partial^2 v}{\partial r^2}  + \frac{\partial^2 v}{\partial z^2} - \frac{v}{r^2} \right)+ \frac{\partial P}{\partial r} = 0 \\
\label{NS3}
e_{NS3} &= \frac{\partial v}{\partial r} + \frac{v}{r} + \frac{\partial u}{\partial z} = 0
\end{align}

\subsubsection{Material Balance}
In dimensionless form, the material balance for each species \textit{i} can be written as:
\begin{align}
\label{MBi}
e_{MB_i} &=  u \frac{\partial C_i}{\partial z} + v \frac{\partial C_i}{\partial r} - \frac{1}{Pe} \cdot \left( \frac{1}{r} \frac{\partial C_i}{\partial r} + \frac{\partial^2 C_i}{\partial r^2} + \frac{\partial^2 C_i}{\partial z^2} \right)+ r_i = 0,
\end{align}
where \( C_i \) is the dimensionless concentration of species \( i \). \( r_i \) is the net rate of consumption for species \( i \). For the second-order reaction, the dimensionless pre-exponential factor,
\[
k_{0,\text{dimensionless}} = \frac{k_0 \cdot D \cdot C_{\text{char}}}{V},
\]
where $C_{\text{char}}$ is the characteristic concentration.


\subsubsection{Energy Balance}
The energy balance equation describes the temperature profile within the reactor. The summation term $\sum_j \Delta H_{r,j} r_j$ accounts for heat release from all chemical reactions occurring within the reactor, where \( \Delta H_{r,j} \) is the enthalpy of reaction for the \( j \)-th reaction, and \( r_j \) is the corresponding reaction rate. The values of enthalpy for each case can be found in Appendix \ref{Appendix}.
In dimensionless form, the energy balance equation can be written as:
\begin{align}
\label{EB1}
e_{EB} = u \frac{\partial T}{\partial z} + v \frac{\partial T}{\partial r} - \frac{1}{Pe_T} \left( \frac{1}{r} \frac{\partial T}{\partial r} + \frac{\partial^2 T}{\partial r^2} + \frac{\partial^2 T}{\partial z^2} \right) +\sum_j \Delta H_{r,j} r_j  = 0
\end{align}
For the second-order reaction, the dimensionless enthalpy of the reaction,
\begin{align*}
\Delta H_{\text{dimensionless}} &= \frac{\Delta H \, C_{\text{char}}}{\rho C_p T_{\text{char}}},
\end{align*}
where $T_{\text{char}}$ is the characteristic temperature, \( \rho \) denotes the fluid density and \( C_p \) is the heat capacity of the fluid.


\subsection{Chemical Reaction Kinetics}
The reaction kinetics and the temperature dependence of the
rate constants are known from the Arrhenius equation \cite{arrhenius1889reaktionsgeschwindigkeit}. \(k_0 \) is the pre-exponential factor, \( E_a \) is the activation energy, \( R \) is the gas constant, and \( T \) is the absolute temperature in Kelvin. The detailed values of \( E_a \) and \(k_0 \) for each case can be found in Appendix \ref{Appendix}.

\begin{equation}
\label{Arr}
k = k_0 \cdot e^{-\frac{E_a}{RT}}
\end{equation}

\section{Methods}
\subsection{Physics-Informed Machine Learning (PIML)}

In this study, we employ Physics-Informed Machine Learning (PIML) to obtain solutions for the system of PDEs described in Section~\ref{Governing_Eqs}. Specifically, we use a parameterized representation model, $\mathcal{M}$, to approximate the desired solutions:
\begin{align*}
    (\bm{u}, p, \bm{C}, T) = \mathcal{M}(\theta, \bm{x}),
\end{align*}
where \( \bm{x} = (r, z) \) represents the cylindrical coordinates, $\bm{u} = \{u, v\}$ are the velocity components in the $z$ and $r$ directions respectively and $p$ is the pressure. $\bm{C} = \{C_i\}_{i=1}^N$ denotes the concentrations of $N$ species, and $T$ is the temperature. Here, $\theta$ represents the trainable parameters of $\mathcal{M}$. 

The choice of the representation model defines the specific PIML variation. For instance, if $\mathcal{M}$ is a multilayer perceptron (MLP), then we recover the PINN formulation introduced in~\cite{raissi2019physics}. Similarly, if $\mathcal{M}$ is a Chebyshev Kolmogorov-Arnold Network (KAN), then the framework is referred to as cPIKANs~\cite{shukla2024comprehensive}. The specific details of the representation models used in this study are provided in Section~\ref{Representation_models}.

\subsubsection{Loss Functions}

To ensure that the obtained predictions are physically meaningful, the model iteratively updates the trainable parameters $\theta$ to minimize a combined loss function ($\mathcal{L}$) that accounts for all required constraints, including the PDEs (i.e., \autoref{NS1}, \ref{NS2}, \ref{NS3}, \ref{MBi}, and \ref{EB1}) and boundary conditions (see Figure~\ref{fig:pfr}(b)). The loss function for the forward problem is:
\begin{align}
    \mathcal{L} = \mathcal{L}_{PDE} + \mathcal{L}_B,
\end{align}
where $\mathcal{L}_{PDE}$ is the loss for PDEs and $\mathcal{L}_B$ is the loss for boundary conditions. Each loss sub-term can be further decomposed as:
\begin{align}
\label{data_eq}\mathcal{L}_G(X_G,\theta) &= \sum_{\alpha} m_\alpha\langle[\lambda_{\alpha,i}e_\alpha(\textbf{x}_i,\theta)]^q\rangle_{i=1}^{n}, 
\end{align}
where $G=\{PDE, B\}$ is an index specifying each loss group. The notation $\langle\cdot\rangle_i$ represents the mean over $n$ training points $\textbf{x}_i = (r_i, z_i)$ in the subset $X_G$, which is sampled uniformly in every iteration from the domain $\Omega_G$. Each loss component controls different variables or quantities, which are identified by the index $\alpha$. In the boundary loss case ($\mathcal{L}_B$), we enforce constraints on the velocities ($u,v$), concentrations ($\{C_i\}_{i=1}^N$), and temperature $T$; therefore $\alpha = \{u,v,p,\{C_i\}_{i=1}^N,T\}$. For the PDE loss ($\mathcal{L}_{PDE}$), we enforce the Navier-Stokes ($NS$), material ($MB_i$), and energy balance ($EB$) equations, therefore $\alpha=\{NS_1,NS_2,NS_3,\{MB_i\}_{i=1}^N, EB\}$. The residual, $  e_\alpha(\textbf{x}_i, \theta) = |\hat{\alpha}(\textbf{x}_i) - \alpha(\textbf{x}_i, \theta)|$, quantifies the mismatch between the predicted value $\alpha(\textbf{x}_i, \theta)$ and the target value $\hat{\alpha}(\textbf{x}_i)$ at the point $\textbf{x}_i \in \Omega_G$. To balance the contributions of different loss terms, we introduce local weights ($\lambda_{\alpha, i}$) to adjust the point-wise residual $e_\alpha(\textbf{x}_i, \theta)$ and global weights ($m_\alpha$) to scale the averaged value of each subcomponent $\alpha$ as shown in Section \ref{rba}.

\paragraph{Inverse Problem.}  
One of the challenges in modeling real data is the presence of unknowns in the PDEs or unspecified boundary conditions. However, the flexibility of PIML allows for solving such problems by incorporating observations inside the domain~\cite{raissi2019physics}. To discover the unknown kinetics parameter, we train our model to learn the activation energy $E_a$ described in \autoref{MBi} and \autoref{EB1}. The combined loss function for this setup is given by:
\begin{align}
    \mathcal{L} = \mathcal{L}_{PDE} + \mathcal{L}_B +  \mathcal{L}_D,
\end{align}
where $\mathcal{L}_D$ represents the data loss. $\mathcal{L}_D$ minimizes the error from sparse concentration and temperature observations inside the domain as shown in Figure~\ref{fig:pfr}(c). Specifically, $\mathcal{L}_D$ also follows Equation \ref{data_eq}, with $\alpha=\{\{C_i\}_{i=1}^N,T\}$. Finally, to infer the hidden parameters, we redefine our parameter set as:
\begin{align*}
    \theta=\{\theta_{\mathcal{M}},E_a\},
\end{align*}
where $\theta_{\mathcal{M}}$ (i.e., the representation model parameters) and $E_a$ (i.e., the inferred activation energy) are concurrently learned by iteratively minimizing the combined loss function. Notably, $E_a$ can be inferred since obtaining an appropriate value is crucial for minimizing the residuals of Equations \ref{MBi} and \ref{EB1}.

\subsubsection{Representation Models}\label{Representation_models}
\paragraph{Multilayer Perceptrons (MLPs).}  

A Multilayer Perceptron (MLP) maps an input \( \textbf{x} = (x_1, x_2, \dots) \) to an output \( y \) through a series of nested transformations:  
\begin{align}
  y(\bm{x}) = \sigma\left(W^{(L)} \sigma\left(W^{(L-1)} \dots \sigma\left(W^{(1)} \textbf{x} + b^{(1)}\right) \dots + b^{(L-1)}\right) + b^{(L)}\right).  
\end{align}
Here, \( L \) denotes the number of layers, while \( W^{(l)} \) and \( b^{(l)} \) are the weights and biases at layer \( l \). Each layer's output serves as the input to the next, progressively transforming the input. With a sufficient number of neurons and an appropriate activation function \( \sigma \), MLPs can approximate any continuous function on compact subsets of \( \mathbb{R}^n \) \cite{hornik1989multilayer}.

\paragraph{Kolmogorov-Arnold Networks (KANs).}  

Kolmogorov-Arnold Networks (KANs) is a recently developed class of representation models based on the Kolmogorov-Arnold representation theorem \cite{liu2024kan}. This theorem states that any continuous multivariate function \( f(\boldsymbol{x}) = f(x_1, x_2, \dots) \) defined on a bounded domain can be expressed as a composition of continuous univariate functions and the addition operation. Inspired by this idea, Liu et al. \cite{liu2024kan} introduced an approximation for \( f(\boldsymbol{x}) \) in the form:
\begin{equation}
  f(\boldsymbol{\zeta}) \approx \sum_{i_{L-1}=1}^{n_{L-1}} \Phi_{L-1,i_L,i_{L-1}} \left( \sum_{i_{L-2}=1}^{n_{L-2}} \cdots \left( \sum_{i_2=1}^{n_2} \Phi_{2,i_3,i_2} \left( \sum_{i_1=1}^{n_1} \Phi_{1,i_2,i_1} \left( \sum_{i_0=1}^{n_0} \phi_{0,i_1,i_0}(\zeta_{i_0}) \right) \right) \right) \cdots \right).
\label{KAN_net}
\end{equation}
The right-hand side of \eqref{KAN_net} defines a \( KAN(\boldsymbol{\zeta}) \), where \( \boldsymbol{\zeta} = (\zeta_1, \zeta_2, \dots) \) represents the input variables, \( L \) denotes the number of layers, and \( \{n_j\}_{j=0}^{L} \) specifies the number of nodes in each layer, \( \phi_{i,j,k} \) and \( \Phi_{i,j,k} \) correspond to inner and outer univariate functions, respectively. Variations in the formulation of \( \phi(x) \) give rise to different KAN architectures. A notable extension, cKANs, was introduced in \cite{shukla2024comprehensive}. These networks employ Chebyshev polynomials as their univariate basis functions, namely:
\begin{align*}
 \phi(\zeta,\theta) = \sum_n c_n T_n(\zeta),
\end{align*}
where \( \theta = \{c_n\} \) represents the trainable parameters, and \( T_n(\zeta) \) denotes the Chebyshev polynomial of order \( n \), recursively defined as:

\[
T_n(\zeta) = 2\zeta T_{n-1}(\zeta) + T_{n-2}(\zeta).
\]
This recursive approach enhances numerical stability by systematically generating higher-order polynomials from lower-order ones. Additionally, the outer functions in cKANs are defined as:
\begin{align*}
 \Phi(\zeta,\theta) = \phi(\tanh(\zeta)).
\end{align*}
As noted in \cite{shukla2024comprehensive}, this transformation ensures that inputs to the univariate functions remain within \([-1,1]\), the domain where Chebyshev polynomials are well-defined. Enforcing this constraint enhances numerical robustness and maintains the accuracy of polynomial expansions. Previous studies have observed that, for particular examples, cKANs exhibit strong resilience to noise \cite{shukla2024comprehensive} and can accelerate convergence when solving complex PDEs \cite{toscano2024kkans}.

\subsection{Additional Physics-Informed Machine Learning (PIML) Enhancements}

\subsubsection{Weight Normalization} 
To accelerate the training process, we adopt weight normalization, a reparameterization technique that decouples the magnitude of weight vectors from their direction. This method has been shown to enhance convergence speed in deep learning models \cite{salimans2016weight,raissi2020hidden,anagnostopoulos2024residual}. Specifically, the output of each neuron is computed as
\begin{align}
y &= \sigma(W \cdot x + b), \\ W &= \frac{g}{\lVert \mathbf{v} \rVert_2} \mathbf{v}, \label{eq_wn} \end{align} \noindent where $y$ denotes the neuron output, $\sigma$ is the activation function, $x$ is the input vector, $W$ is the weight vector, and $b$ is the bias. As shown in equation~\ref{eq_wn}, the weight vector $W$ is reparameterized in terms of two new trainable parameters: a direction vector $\mathbf{v}$ and a scalar scale parameter $g$. This formulation ensures that $\lVert W \rVert_2 = g$, effectively separating the weight's magnitude from its direction. Such decoupling promotes faster and more stable convergence while introducing negligible computational overhead~\cite{salimans2016weight}.

\subsubsection{Output Transformation}
\paragraph{Exact Imposition of Dirichlet Boundary Conditions.}
The inexact enforcement of boundary conditions can negatively affect the performance and training stability of neural networks \cite{dong2021method,wang2021understanding,chen2020comparison}. 
In our model, we enforce exact Dirichlet boundary conditions for the velocity components $u$ and $v$ at the wall within the Navier–Stokes sub-network, following the method proposed by Sukumar et al. \cite{sukumar2022exact}. In their work, Dirichlet boundary conditions in PINNs are imposed using approximate distance functions (ADF). The constrained formulation for the Dirichlet boundary conditions is given by Equation \ref{ADF}: 

\begin{equation}
 u(\mathbf{x})=g(\mathbf{x})+\varphi(\mathbf{x})u_{NS}(\mathbf{x}),
 \label{ADF}
\end{equation}
where $g(\mathbf{x})$ is a function that satisfies the boundary conditions for $u$, while $u_{NS}(\mathbf{x})$ is the output of the Navier–Stokes sub-network, and $\varphi(\mathbf{x})$ is a composite distance function that is zero along the boundaries.

\paragraph{Constrained Domain.}
Constrained domain refers to the restriction of the neural network's output to lie within a predetermined range, achieved by constructing the mapping \(\Gamma: \mathbb{R}^p \rightarrow \mathbb{R}^q\). Both nondimensionalization and constrained domain methods have been successfully employed in various multiscale problems \cite{anagnostopoulos2024residual,mao2020physics,zapf2022investigating}. Given the inherently multiscale and multiphysics nature of reactor design, we nondimensionalize the governing equations and construct \(\Gamma\) to capture the specific characteristics of the problem. This constrained domain approach enables the incorporation of prior knowledge regarding the expected ranges for the concentration \(C_i\) and the temperature \(T\). However, as demonstrated in our ablation study, prior knowledge is not essential for the model's overall success.

\subsubsection{Weighting}
\paragraph{Local Weights.}\label{rba}
Due to the multiphysics and multiscale nature of reactor design, concurrently minimizing multiple loss functions poses a significant challenge. To address this, sampling \cite{wu2023comprehensive} and local weighting strategies \cite{mcclenny2023self} are typically employed. In this study, we utilize local multipliers residual-based attention (RBA) weights \cite{anagnostopoulos2024residual}, which balance the contributions of individual training points within each loss term and induce residual homogeneity \cite{anagnostopoulos2024learning}. The update rule for the RBA weights for each training point \(i\) at iteration \(k\) is given by:

\begin{equation}
  \lambda_i^{k+1} \leftarrow \gamma\lambda_i^{k} + \eta^* \frac{\left| e_i \right|}{\lVert e \rVert_{\infty}}, \quad i \in \{0, 1, \dots, N\},
  \label{Update_RBA}
\end{equation}

\noindent where \(N\) denotes the total number of training points, \(e_i\) is the residual associated with the loss term at point \(i\),  \(\gamma\) is a memory (decay) rate, and \(\eta^*\) is a learning rate. This convergent linear homogeneous recurrence relation constrains the RBA weights to lie within the interval \([0,\eta^*/(1-\gamma)]\).

\paragraph{Global Weights.} 
The global weights adjust and balance the contributions of each term in the loss function, ensuring that all constraints are satisfied. These weights may be defined as fixed, as in the original PIML framework~\cite{raissi2019physics, raissi2020hidden}, or implemented dynamically~\cite{wang2021understanding, xiang2022self, liu2021dual, basir2022investigating, wang2022respecting, wang2022and}. In both configurations, global weights are crucial to the efficacy of PIML models, and ongoing research continues to refine their implementation to enhance overall performance. In this study, fixed global weights are applied at each training stage; however, due to the multi-stage nature of our model, these weights evolve dynamically in an iteration-based manner.

\subsubsection{Sequential Training}
Incorporating sequential training into physics-informed machine learning has been demonstrated to improve predictive accuracy by incrementally increasing the complexity of the problem. This strategy has been successfully applied to forward systems of PDEs \cite{krishnapriyan2021characterizing,zhang2024meshless,wang2023expert,wang2024piratenets} as well as inverse problems \cite{ahmadi2024ai,daryakenari2025cminns,toscano2024invivo,toscano2024inferring}. Its implementation is tailored to the specific problem context. For example, Wang et al. \cite{wang2023expert} addressed the 2D Navier–Stokes equations at high Reynolds numbers by incrementally increasing the Reynolds number during training. In inverse problems involving the inference of hidden fields, Toscano et al. \cite{toscano2024inferring} employed a sequential approach that first fits sparse experimental data and boundary conditions, then learns a theoretical representation of the hidden field, and finally integrates the full physics. In the context of reactor design, which encompasses transport, kinetics, material balance, and energy balance, we initially focus on the Navier–Stokes equations to capture fluid velocities in the plug flow reactor, and subsequently increase the system complexity.

\subsection{Flow, Material and Energy Nets (FMEnets)}

\begin{figure}[htbp]
    \centering
    \includegraphics[scale = 0.6]{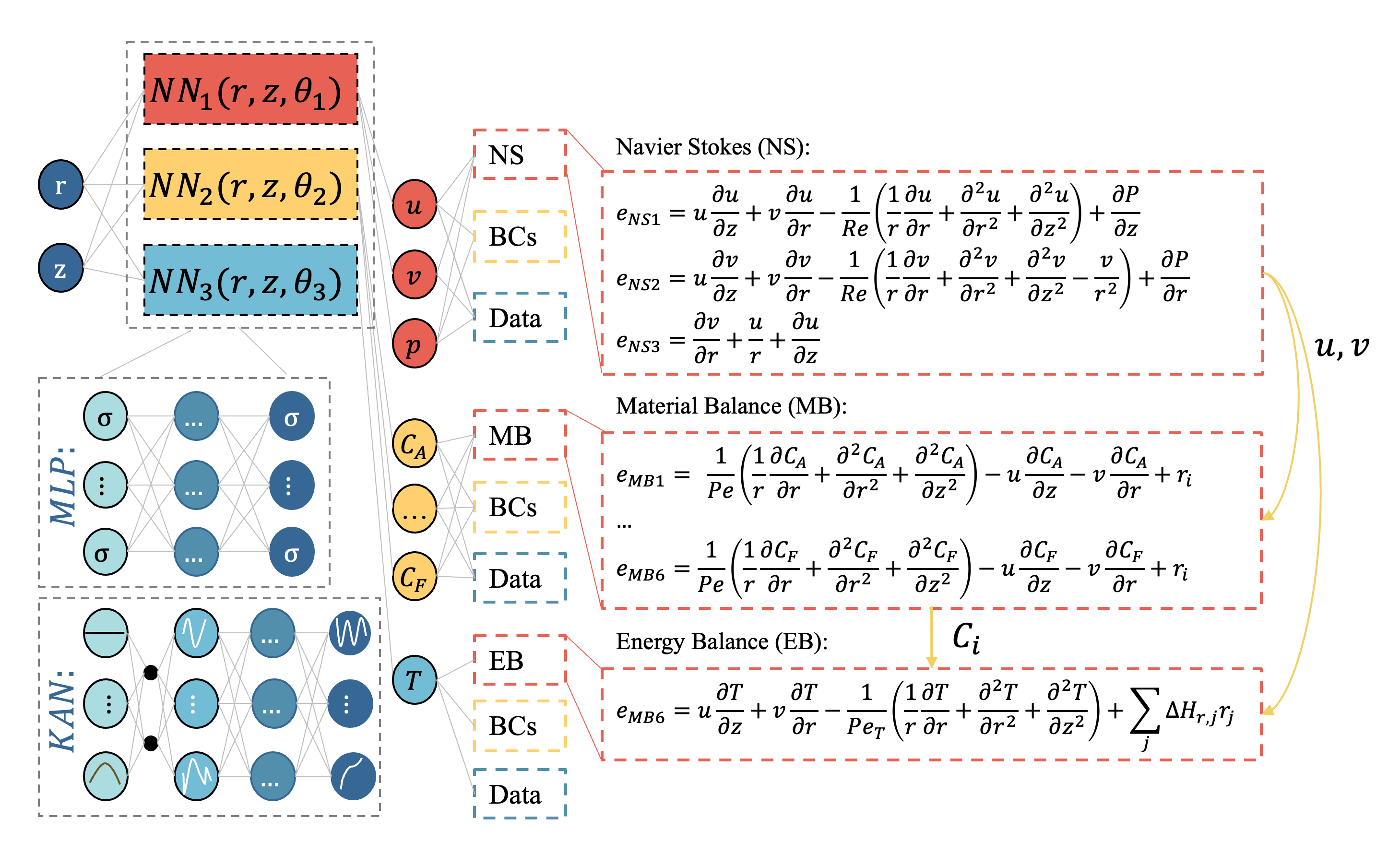}
    \caption{Schematic of the FMEnets architecture for solving the forward problem. Three neural networks ($NN_1, NN_2, NN_3$) each enforce different governing equations, including Navier–Stokes (NS), material balance (MB), and energy balance (EB), along with boundary conditions and inlet/outlet experimental data. $NN_1$ first predicts the velocity field ($u, v$), which then feeds into $NN_2$ and $NN_3$ to learn species concentrations and temperature, respectively. Each sub-network ($NN_1, NN_2, NN_3$) can be implemented as either a KAN or an MLP.}
    \label{fig:nn1}
\end{figure}

The FMEnets architecture is a physics-informed machine learning framework designed to address forward and inverse problems in reactive plug flow systems by integrating governing equations, boundary conditions, and experimental data. The architecture consists of three neural networks, \(NN_1\), \(NN_2\), and \(NN_3\), each responsible for enforcing specific governing equations, as illustrated in Figure~\ref{fig:nn1}. The first network, NN1, predicts the velocity field \((u,v)\) by solving the Navier–Stokes equations. Building on this, \(NN_2\) employs the velocity information to resolve the material balance equations and determine the species concentrations \(C_i\). \(NN_3\) utilizes the velocity field and species concentrations to enforce the energy balance (EB) equations and determine the temperature field. 

Each sub-network can be implemented using either an MLP or a KAN, leading to two subcategories of FMEnets: \textbf{FME-PINNs}, which employs physics-informed neural networks (PINNs) with MLPs, and \textbf{FME-KANs}, which utilizes KANs. To enhance convergence speed, weight normalization can be incorporated, while output transformations are applied when the expected ranges of concentrations \(C_i\) and temperature \(T\) are known. Additionally, a local weighting scheme ensures balanced contributions from different training points, while a global weighting strategy coupled with sequential training—facilitates a progressive learning process. The training procedure follows a two-step approach: first, FMEnets is trained solely on the constraints of \(NN_1\), incorporating the NS equations, boundary conditions, and data for \(C_i\) at the inlet and outlet. After 30,000 iterations, the parameters of \(NN_1\) are frozen, and training proceeds with the MB and EB equations, using a global weight ratio of 10:1 to account for the significantly larger number of equations in \(NN_2\) compared to \(NN_3\). In this second phase, only the parameters of MB and EB are updated.  

FMEnets combines multi-scale data with governing physical principles to describe complex reacting flow systems. Its flexible network design and training methods is promising for various multiphysics modeling tasks in chemical engineering.

\subsection{Inverse FMEnets for Inferring Unknown Kinetic Parameters}
\label{sec: inverse}
In practical applications, direct measurement of concentration and temperature profiles inside the reactor is often infeasible, making the inverse problem ill-posed \cite{rathore2024challenges}. The inverse FMEnets framework extends the FMEnets architecture to infer unknown kinetic parameters in reactive flow systems by incorporating additional experimental data within the reactor. This study represents the first attempt to recover kinetic parameters in reactive flow systems using the physics-informed machine learning model. To overcome the lack of interior measurements, multi-residence-time experimental data are incorporated. These data are obtained by considering reactors of varying lengths—specifically, a full-length reactor, a half-length reactor, and a quarter-length reactor. As illustrated in Figure~\ref{fig:pfr}(c), we use multi-residence-time experimental data at the outlet as a surrogate for interior measurements. In this study, we use simulated data instead of  experimental data to conduct a proof-of-concept that demonstrates the viability of our FMEnets. By leveraging sparse observations at the inlet and outlet, along with measurements at intermediate positions (e.g., at half and quarter reactor lengths), the inverse FMEnets enables the inference of unknown kinetic parameters, such as the activation energy \(E_a\), while simultaneously predicting species concentrations and temperature profiles.

\begin{figure}[htbp]
    \centering
    \includegraphics[scale = 0.6]{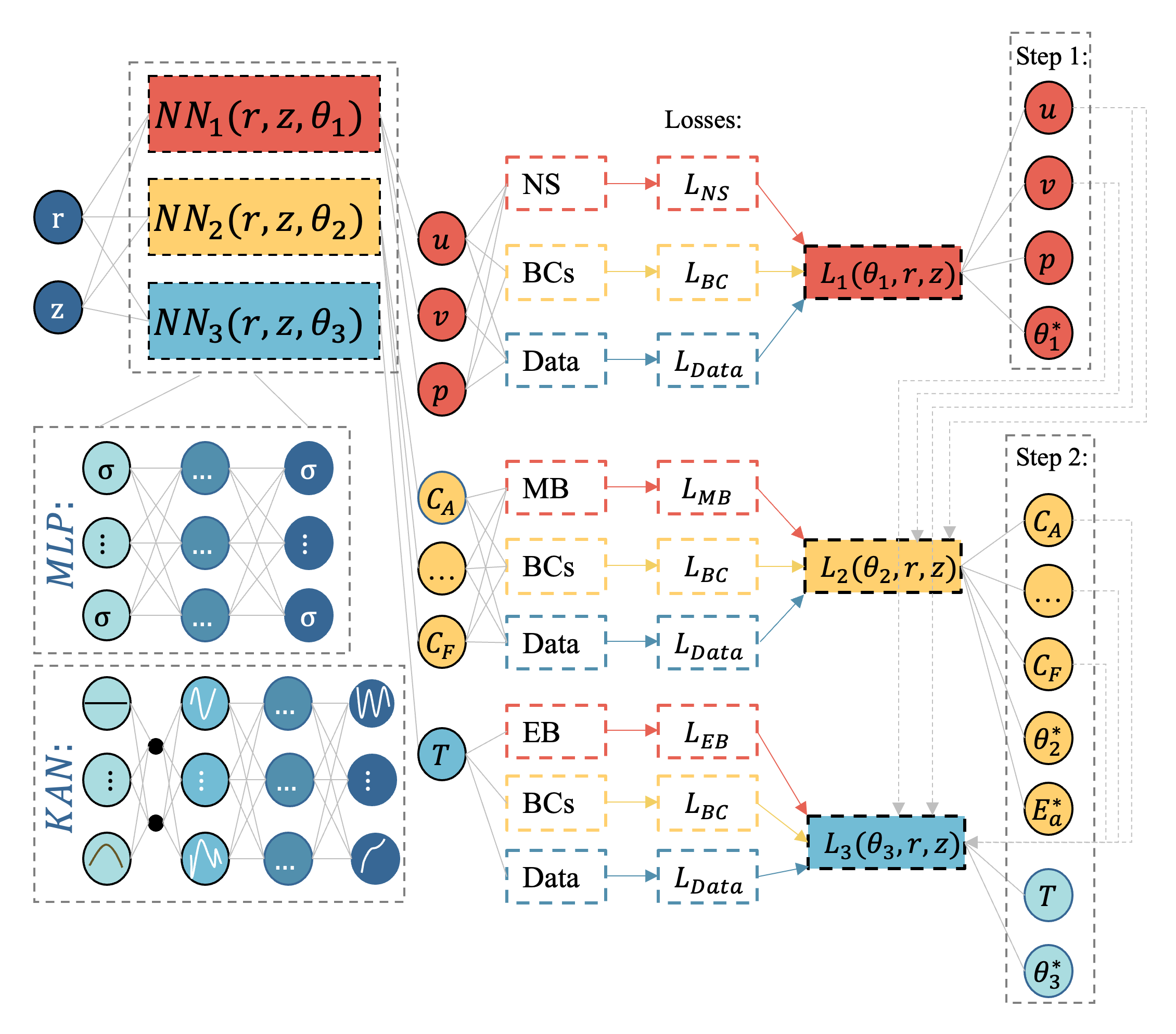}
    \caption{Schematic of the FMEnets architecture for the inverse problem with a two‐step procedure. Step 1: We train $NN_1$ to enforce the Navier–Stokes (NS) equations and predict the velocity field ($u,v$). Step 2: $NN_2$ and $NN_3$ employ those velocity predictions to solve the material (MB) and energy balance (EB) equations with boundary conditions and multi‐residence‐time experimental data while inferring the unknown activation energy $E_a$. Each sub-network ($NN_1, NN_2, NN_3$) can be implemented as either a KAN or an MLP.}
    \label{fig:nn2}
\end{figure}

Similar to forward FMEnets, the inverse FMEnets also employs a two-step training procedure as depicted in Figure~\ref{fig:nn2}. Step 1 trains \(NN_1\) to enforce the Navier–Stokes equations, boundary conditions, and available experimental data to predict the velocity field \((u, v)\). In Step 2, the predicted velocity field from \(NN_1\) is used by \(NN_2\) and \(NN_3\) to solve the material balance and energy balance equations, respectively. The framework supports both MLPs and KANs. Unlike the forward FMEnets, the sequential training strategy for the inverse problem is modified to account for the presence of unknown kinetic parameters in the MB and EB equations. Since these equations are less reliable until a physically consistent \(E_a\) is identified, the model prioritizes learning the kinetic parameters before refining concentration and temperature predictions. The Navier–Stokes equations are not influenced by the unknown kinetic parameters, enabling Step 1 to follow the same procedure as in the forward FMEnets. During Step 2, the global weights assigned to the MB and EB equations gradually increase as training progresses, thus it enables the model to first identify an appropriate \(E_a\) based on experimental data. After identifying \(E_a\), the model focuses on optimizing the predictions of concentration and temperature fields. By integrating multi-residence-time experimental data with a physics-informed deep learning approach, the inverse FMEnets provide a robust and generalizable framework for kinetic parameter inference in complex reacting flow systems.

\subsection{Data Generation}

To validate the results obtained from the PINN against numerical solutions, we solve the system of equations presented in \autoref{NS1}-\autoref{NS3}, \autoref{MBi}, and \autoref{EB1} using the finite element method (FEM). Here, we selected FEM  to address two critical challenges in this system: ensuring accuracy and preventing singularities at \( r = 0 \) through the careful formulation of the numerical scheme. To achieve this, we derive the variational formulation of \autoref{MBi} by multiplying the test function \( \bm{\phi} \in \hat{V} = \left\{ \phi \in H^1(\Omega) | \phi = 0 \, \text{on} \, \partial \Omega \right\} \), where \( \hat{V} \) denotes the space of test functions, and then integrate over the domain \( \Omega \). Here, \( H^1(\Omega) \) is the Sobolev space containing finite integral of $\phi^2$ and $|\nabla \phi|^2$. 

\begin{align}\label{eq:weak_formulation}
\int_{\Omega} \left(\bm{u} \cdot \nabla C\right)\bm{\phi} {~\rm d\Omega} - \frac{1}{\rm P_e} \int_{\Omega} \Delta C\bm{\phi} {~\rm d\Omega} + \int_{\Omega} \frac{1}{r}\frac{\partial C}{\partial r}\bm{\phi} {~\rm d\Omega} + \int_{\Omega} r \bm{\phi} {~\rm d\Omega} = 0.
\end{align}
We now perform integration by parts to the third term of \autoref{eq:weak_formulation} and obtain
\begin{align}\label{eq:var_prob}
\int_{\Omega} \left(\bm{u} \cdot \nabla C\right)\bm{\phi} {~\rm d\Omega} - \frac{1}{\rm P_e} \left[ \int_{\rm d \Omega_{N}} \bm{\phi } \nabla C \cdot \hat{\bm{n}}~{\rm d s} -\int_{\Omega} \nabla C \nabla \bm{\phi}~{\rm d\Omega} \right] + \int_{\Omega} \frac{1}{r}\frac{\partial C_h}{\partial r}\bm{\phi} {~\rm d\Omega} + \int_{\Omega} r \bm{\phi}~{\rm d \Omega} = 0,
\end{align}
where ${\rm ds}$  denote the differential element for integration over $\partial \Omega$, the Neumann boundary of domain $\Omega$.

The variational problem in \autoref{eq:var_prob} is continuous and seeks the solution \( C \) in the trial space \( V = \left\{ \phi \in H^1(\Omega) \mid \phi = C_D \text{ on } \partial \Omega \right\} \), where \( C_D \) denotes the Dirichlet boundary condition on \( C \). We will now express \autoref{eq:var_prob} in its discretized form.

\begin{align}\label{eq:discrete}
\int_{\Omega} \left(\bm{u} \cdot \nabla C_h\right)\bm{\phi} {~\rm d\Omega} - \frac{1}{\rm P_e} \left[ \int_{\rm d \Omega_{N}} \bm{\phi } \nabla C_h \cdot \hat{\bm{n}}~{\rm d s} -\int_{\Omega} \nabla C_h \nabla \bm{\phi}~{\rm d\Omega} \right] + \int_{\Omega} \frac{1}{r}\frac{\partial C_h}{\partial r}\bm{\phi} {~\rm d\Omega} + \int_{\Omega} r_h \bm{\phi}~{\rm d \Omega} = 0,
\end{align}
where $C_h \in V_h \subset V$ and $\bm{\phi} \in \hat{V}_h \subset \hat{V}$.

\begin{figure}[htbp]
    \centering
    \includegraphics[scale = 0.6]{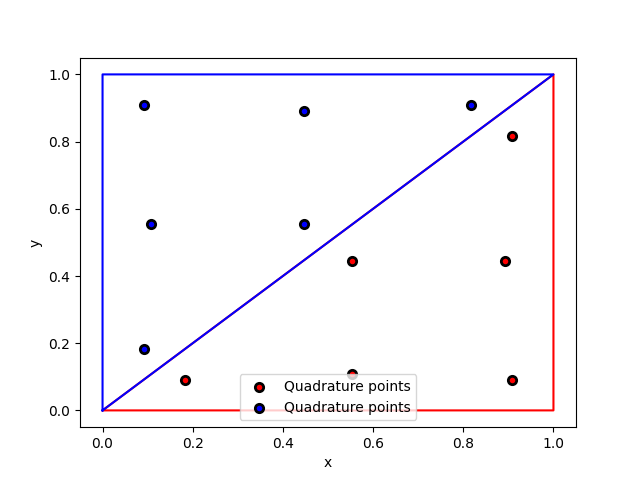}
    \caption{Distribution of 6-point Gauss quadrature points on a reference triangle and exact integration for polynomial degree of 4.}
    \label{fig:gauss_quad}
\end{figure}

To numerically compute the integrals in \autoref{eq:var_prob}, we employ Gauss quadrature points, which helps avoid the evaluation of the singular term (the third term in \autoref{eq:discrete}) at \( r = 0 \). To demonstrate this, we display the distribution of 6-point Gauss quadratures in Figure \ref{fig:gauss_quad}. In Figure \ref{fig:gauss_quad}, we show the distribution of Gauss quadrature points in a domain discretized with two triangular elements, which is exact for polynomials of degree 4. The quadrature points do not coincide with \( r = 0 \), thus mitigating instability. 

For implementing the numerical scheme in \autoref{eq:discrete}, we used the FEniCS codebase \cite{alnaes2015fem} to code the variational form \autoref{eq:discrete} along with the appropriate boundary conditions. The domain was discretized using 120,000 uniform triangular elements with a polynomial degree 2. The reaction term requires solving a nonlinear system of equations, and for this, we use the SNES solver \cite{petsc-user-ref} in combination with the MUMPS-based linear solver \cite{amestoy2001mumps}. The absolute and relative tolerance for the solver's convergence is set to \( 10^{-10} \).

\section{Results}
\subsection{Forward FME-PINNs}
For the forward FME-PINNs, we assume sparse observations of species concentrations are available at the inlet and outlet flows. Since the plug flow reactor is immersed in an isothermal bath, the wall temperature is known. The detailed values for Reynolds number (\(Re\)), Péclet numbers (\(Pe, Pe_T\)), reaction rate constant (\(k_0\)), activation energy (\(E_a\)), reaction enthalpy (\(H_{r,j}\)), and other relevant physical and chemical properties are provided in Appendix \ref{Appendix}.  

\begin{figure}[htbp]
    \centering
    \includegraphics[scale = 0.35]{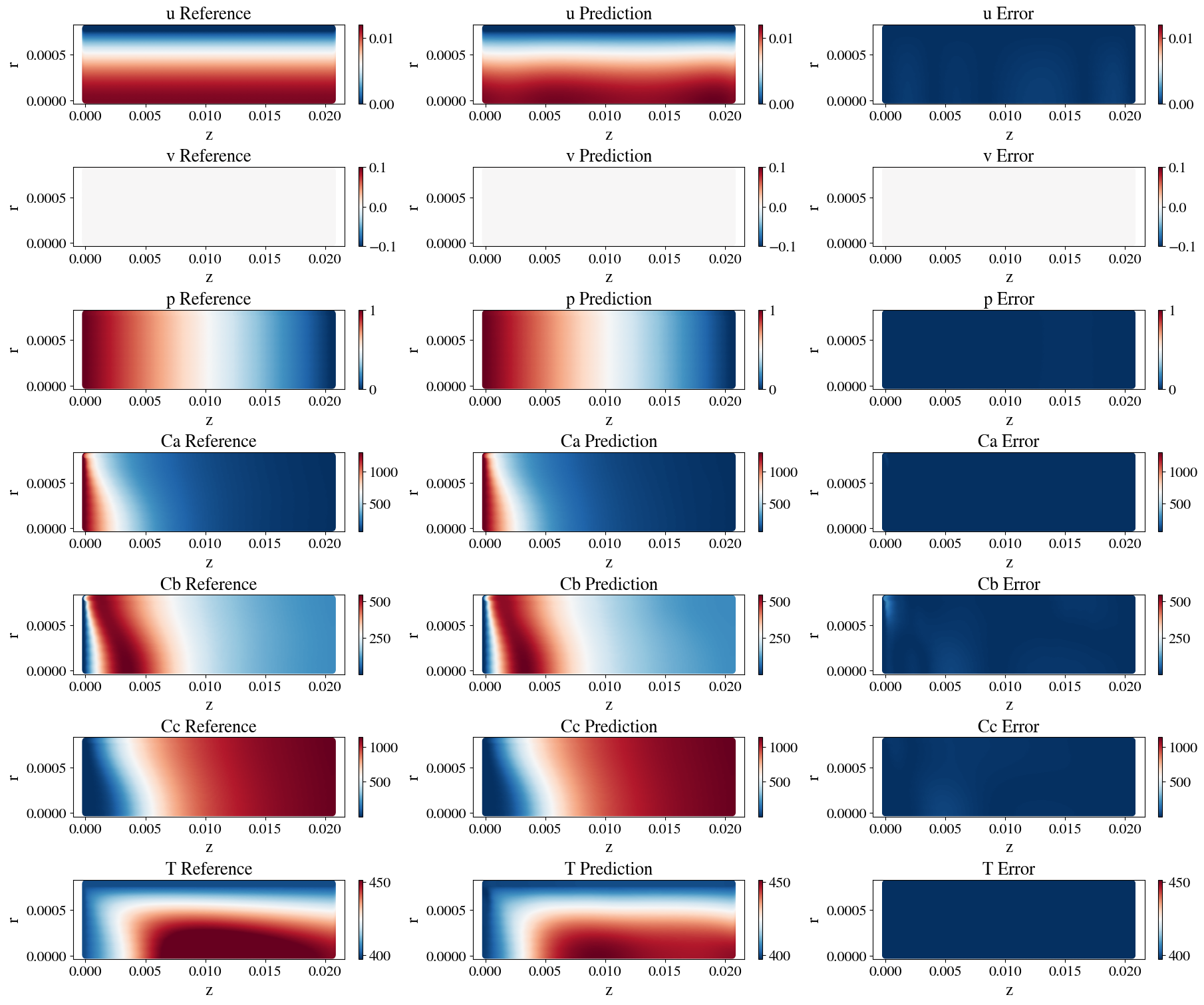}
    \caption{Visualization of the accuracy of the FME-PINNs against Finite Element Method (FEM) simulations for Case\,2: FEM (left column), FME-PINNs (middle column), and the absolute error (right column) for each variable, including velocity components (\(u, v\)) in $m/s$, pressure (\(p\)) in $Pa$, species concentrations (\(C_a, C_b, C_c\)) in $mol/m^3$, and temperature (\(T\)) in $K$.}
    \label{fig:case2_forward}
\end{figure}

\begin{figure}[htbp]
    \centering
    \includegraphics[scale = 0.35]{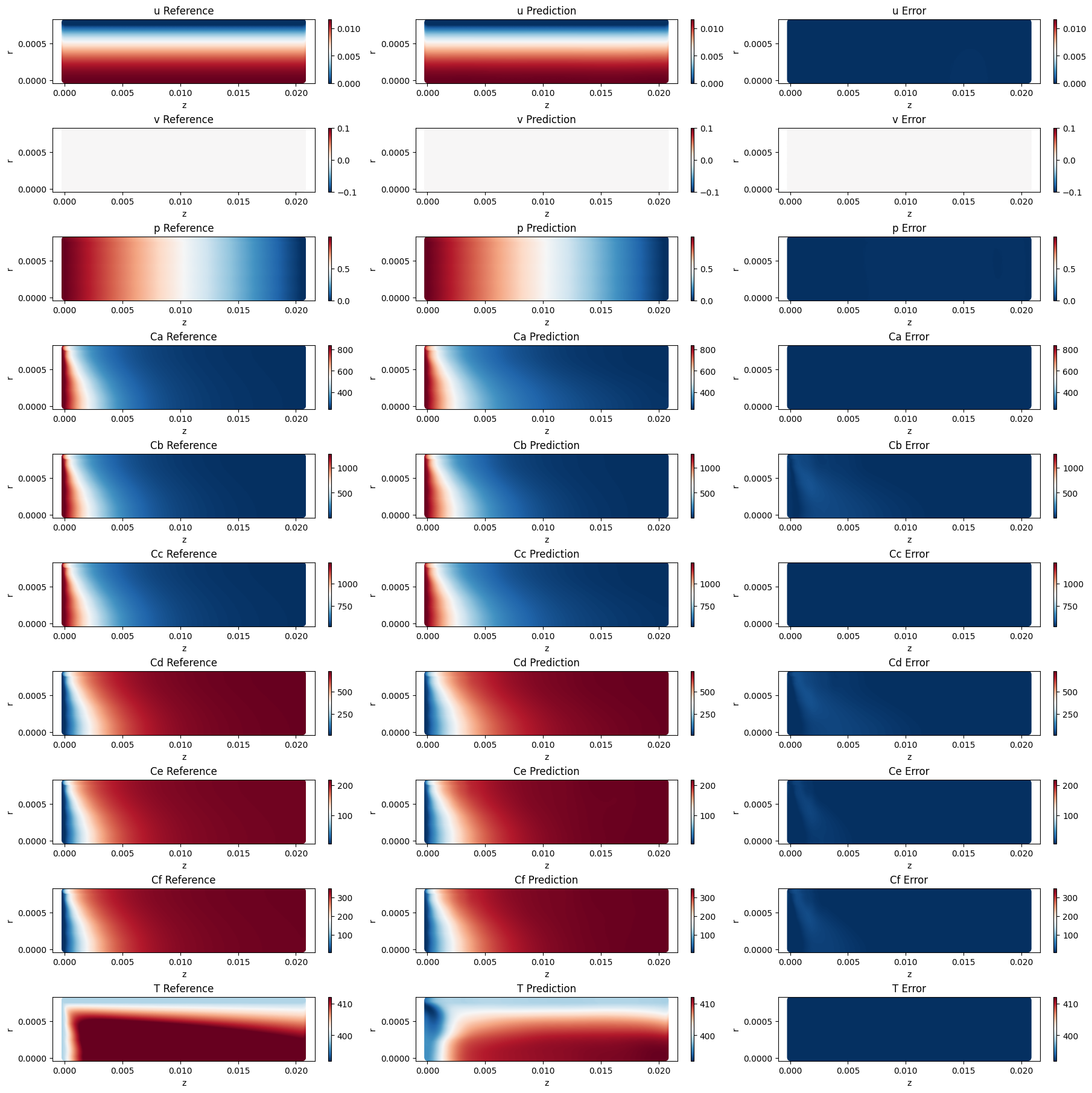}
    \caption{Visualization of the accuracy of the FME-PINNs against Finite Element Method (FEM) simulations for Case\,3: FEM (left column), FME-PINNs (middle column), and the absolute error (right column) for each variable, including velocity components (\(u, v\)) $m/s$, pressure (\(p\)) in $Pa$, species concentrations (\(C_a, C_b, C_c, C_d, C_e, C_f\)) in $mol/m^3$, and temperature (\(T\)) in $K$.}
    \label{fig:case3_forward}
\end{figure}

Figures~\ref{fig:case1_forward},~\ref{fig:case2_forward} and~\ref{fig:case3_forward} present a comparative analysis of FME-PINNs predictions against Finite Element Method (FEM) simulations for velocity components (\(u, v\)), pressure (\(p\)), species concentrations (\(C_a, C_b, C_c\)), and temperature (\(T\)) for two-component and three-component reaction systems, respectively. In these figures, the left column illustrates the FEM solutions (reference), the middle column displays the corresponding FME-PINNs predictions, and the right column provides the absolute error distributions. Figures~\ref{fig:case1_forward},~\ref{fig:case2_forward} and~\ref{fig:case3_forward} demonstrate that FME-PINNs accurately capture the flow, concentrations, and temperature profile patterns for all three cases. The absolute error distributions indicate that the highest errors arise in regions with sharp concentration gradients and temperature variations due to the complex interactions between advection, diffusion, and reaction kinetics. The overall predictive performance of FME-PINNs remains robust. The relative \(L^2\) errors for each component are summarized in Table \ref{tab:cases}. These results show that the FMEnets can be a computationally efficient alternative to traditional numerical solvers for modeling coupled transport and reaction phenomena in reactive flow systems.

\begin{table}[h]
\centering
\begin{tabular}{|>{\centering\arraybackslash}m{4cm}
                |>{\centering\arraybackslash}m{2cm}
                |>{\centering\arraybackslash}m{2cm}
                |>{\centering\arraybackslash}m{2cm}
                |>{\centering\arraybackslash}m{2cm}|}
\hline
\textbf{$L^2$ relative error} & \textbf{Variables} & \textbf{Case 1} & \textbf{Case 2} & \textbf{Case 3} \\ \hline
\label{tab:forward}
\multirow{7}{*}{FME-PINNs}
 & $u$ & 0.736\% & 0.584\% & 0.771\% \\
 & $v$ & 0.000\% & 0.000\% & 0.000\% \\
 & $p$ &  0.455\% & 0.274\% & 0.594\% \\
 & $C_a$ & 1.497\% & 2.334\% & 6.051\% \\
 & $C_b$ & 0.632\% & 2.734\% & 7.952\% \\
 & $C_c$ & -- & 0.991\% & 3.068\% \\
 & $C_d$ & -- & -- & 2.209\% \\
 & $C_e$ & -- & -- & 2.127\% \\
 & $C_f$ & -- & -- & 2.171\% \\
 & T     & 0.965\% & 1.206\% & 1.660\% \\
\hline
\end{tabular}
\caption{\(L^2\) relative errors (in percent) for forward FME-PINNs in three scenarios (\hyperref[para: case1]{\textcolor{red}{Case 1: Two-Component Reaction System}}, \hyperref[para: case2]{\textcolor{red}{Case 2: Three-Component Reaction System}}, and \hyperref[para: case3]{\textcolor{red}{Case 3: Six-Component Reaction System}}) across velocity components (\(u, v\)), pressure (\(p\)), the species concentrations and temperature \(T\). The dash (\(-\)) indicates that \(\mathrm{C_i}\) is not modeled in Case\,1 and Case\,2.}
\label{tab:cases}
\end{table}

\begin{figure}[htbp]
    \centering
    \includegraphics[scale = 0.35]{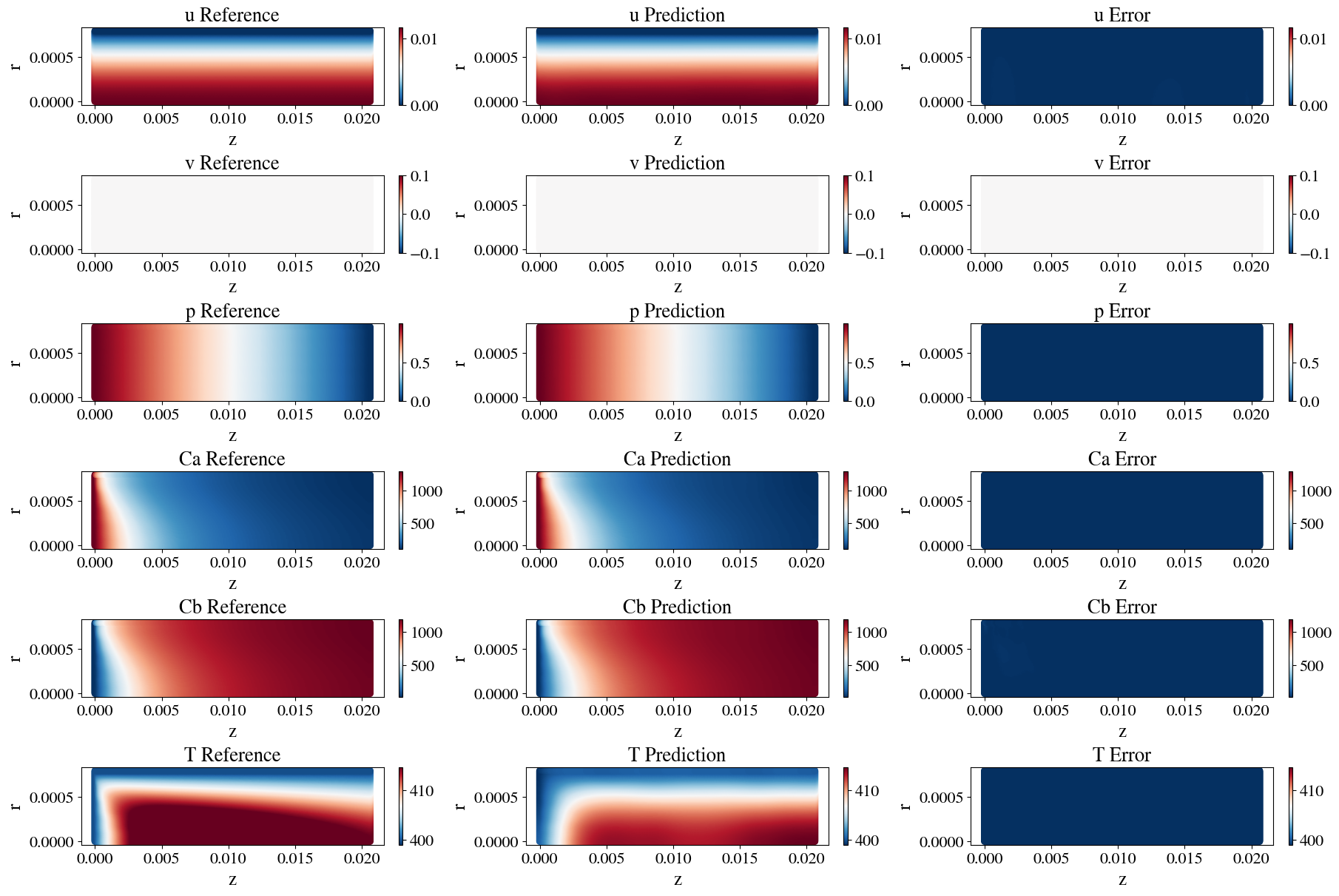}
    \caption{Visualization of the accuracy of the FME-PINNs against Finite Element Method (FEM) simulations for Case\,1: FEM (left column), FME-PINNs (middle column), and the absolute error (right column) for each variable, including velocity components (\(u, v\)) in $m/s$, pressure (\(p\)) in $Pa$, species concentrations (\(C_a, C_b\)) in $mol/m^3$, and temperature (\(T\)) in $K$.}
    \label{fig:case1_forward}
\end{figure}

\subsection{Forward FME-KANs}
In real-world scenarios, the experimental data for the inlet and outlet flow in a plug flow reactor are often noisy due to two primary factors. First, because data for the inlet and outlet are not recorded precisely at the reactor's start and end, system downtime introduces noise in concentration measurements. Second, we often extract concentration measurements for organic products from spectral data, which are calculated as the percentage area under the curve for each species. The process of translating these spectral readings into concentration values introduces additional sources of error.

\begin{figure}[htbp]
    \centering
    \includegraphics[scale = 0.95]{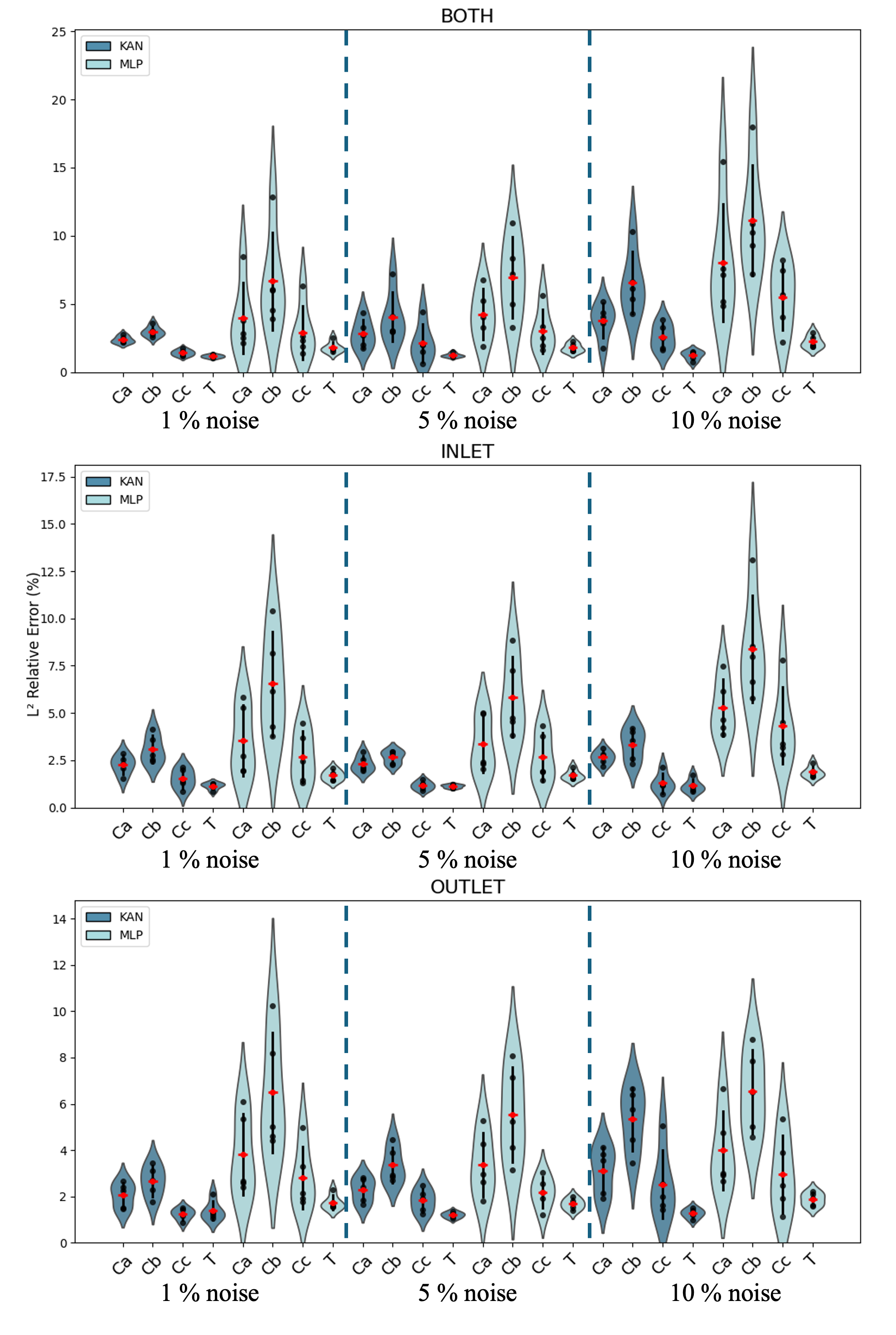}
    \caption{Violin plots of the $L^2$ relative error for FME-KANs (dark blue) and FME-PINNs (light blue) across three noise levels (1\%, 5\%, and 10\%) at different locations (inlet flow (INLET), outlet flow (OUTLET), and both inlet and outlet flow (BOTH)). Each violin illustrates the error distribution for $C_a$, $C_b$, $C_c$, and $T$. The red line marks the mean, the black line indicates mean $\pm$ standard deviation, and black dots represent individual data points.}
    \label{fig:noise_plot}
\end{figure}

Given these challenges, we aim to develop a model for a plug flow reactor that remains robust with noisy data. In this study, we introduce noise at different locations in the reactor: inlet, outlet, and both inlet and outlet. We then evaluate the performance of FME-KANs under three noise levels: 1\%, 5\%, and 10\%. To control the effects of randomness, each scenario is run across five different random seeds. Figure~\ref{fig:noise_plot} presents violin plots of the $L^2$ relative error for FME-KANs and FME-PINNs across varying noise levels and measurement locations, including inlet flow (INLET), outlet flow (OUTLET), and both inlet and outlet flow (BOTH). In general, increasing noise levels result in greater error variability for both architectures. At lower noise levels (1\%), FME-KANs demonstrate relatively small errors with tightly clustered distributions, with all trials yielding errors below 5\%. The average errors for $C_a$ and $C_b$ remain below 3\%, while those for $C_c$ and $T$ are below 1.6\%. However, although FME-PINNs generally perform only slightly worse than FME-KANs, the presence of an outlier with an error exceeding 10\% for $C_b$ skews the overall distribution, indicating increased sensitivity to noise. As noise increases to 5\%, the error distribution broadens significantly for FME-PINNs but only marginally for FME-KANs, demonstrating the robustness of FME-KANs. Even at 10\% noise, FME-KANs maintain low errors for all variables. The highest average error for FME-PINNs occurs for $C_b$, reaching 7\% to 11\% under 10\% noise conditions, whereas the average error for $C_b$ in FME-KANs remains within 3\% to 5\%. These results suggest that FME-KANs are more resilient to noise, making them a strong candidate for handling experimental data with uncertainty and measurement errors.

\subsection{Inverse FME-PINNs}
Table~\ref{tab: inverse} presents the \(L^2\) relative errors for inverse FME-PINNs across two scenarios: a two-component reaction system and a three-component reaction system. The results demonstrate the performance of FME-PINNs in predicting species concentrations (\(C_a, C_b, C_c\)), temperature (\(T\)), and the unknown kinetic parameter (\(E_a\)). In the two-component system, FME-PINNs achieve a relative error of 6.181\% for \(C_a\) and 1.920\% for \(C_b\). The temperature prediction error is 2.182\%, and the inferred \(E_a\) has an error of 1.198\%. In the three-component system, the errors for \(C_a\), \(C_b\), and \(C_c\) are 7.920\%, 8.407\%, and 3.647\%, respectively. The activation energy prediction error increases to 2.112\%. 

The accuracy of concentration and temperature predictions is highly dependent on the precision of the inferred \(E_a\), as the reaction term \(r_j\) is exponentially sensitive to \(E_a\). The impact of \(E_a\) on temperature prediction is more substantial than on species concentrations because the energy balance equation accounts for the cumulative heat release or absorption from all reactions. This effect is evident when comparing inverse FME-PINNs to the forward FME-PINNs with a known \(E_a\). With unknown \(E_a\), we find that the prediction accuracy for temperature decreases in both cases, with errors of 2.182\% and 1.685\% for the two-component and three-component systems, respectively. However, the overall performance of FME-PINNs remains strong, as the inferred activation energy \(E_a\) is predicted with reasonable accuracy in both scenarios. The error is slightly higher in the three-component system than in the two-component system. This increment aligns with our expectations since the three-component system involves two simultaneous reactions. Each reaction has an unknown \(E_a\), making the optimization process more challenging. Our results indicate that the inverse FME-PINNs can effectively infer kinetic parameters even with limited data, as described in Section \ref{sec: inverse}.

\begin{table}[h]
\centering
\begin{tabular}{|>{\centering\arraybackslash}m{4cm}|>{\centering\arraybackslash}m{2cm}|>{\centering\arraybackslash}m{2cm}|>{\centering\arraybackslash}m{2cm}|}
\hline
\textbf{$L^2$ relative error} & \textbf{Variables} & \textbf{Case 1} & \textbf{Case 2} \\ \hline
\multirow{4}{*}{FME-PINNs} & $C_a$ & 6.181\% & 7.920\%  \\
 & $C_b$ & 1.920\% & 8.407\% \\
 & $C_c$ & -- & 3.647\% \\
 & T & 2.182\% & 1.685\% \\ 
 & $E_a$ & 1.198\% & 2.112\% \\ \hline
\end{tabular}
\caption{Comparison of the \(L^2\) relative errors (in percent) for inverse FME-PINNs in two scenarios \hyperref[para: case1]{\textcolor{red}{Case 1}} and \hyperref[para: case2]{\textcolor{red}{Case 2}} across the species concentrations and temperature \(T\), and kinetic parameter \(E_a\). The dash (\(-\)) indicates that \(\mathrm{C_c}\) was not modeled in Case\,1.}
\label{tab: inverse}
\end{table}

\subsection{Inverse FME-KANs}

Table~\ref{tab: inverse_kan} presents a comparison of the \(L^2\) relative errors obtained using FME-PINNs and FME-KANs across species concentrations (\(C_a\) and \(C_b\)), temperature (\(T\)), and the inferred kinetic parameter (\(E_a\)). Both architectures exhibit comparable overall accuracy. 
Specifically, FME-PINNs have marginally lower errors in species concentration predictions, reporting 6.181\% for \(C_a\) and 1.920\% for \(C_b\). 

FME-KANs have errors of 7.950\% and 3.163\% for \(C_a\) and \(C_b\), respectively. However, FME-KANs demonstrate slightly better performance in temperature prediction with an error of 1.731\%. For inferring the unknown activation energy \(E_a\), FME-PINNs have an error of 1.198\%, which is slightly lower than that of FME-KANs. To ensure fair comparisons between the two models, we configure FME-KANs and FME-PINNs with a similar number of parameters, approximately 13,000, and under identical hyperparameters. 

Furthermore, regarding computational efficiency, FME-KANs outperform FME-PINNs, exhibiting average iteration times of 1.0 seconds relative to 1.3 seconds on an NVIDIA GeForce RTX 3090 GPU. These findings indicate that while both architectures effectively solve inverse problems, FME-KANs provide a slight advantage in computational speed without sacrificing accuracy.

\begin{table}[h]
\centering
\begin{tabular}{|>{\centering\arraybackslash}m{4cm}|>{\centering\arraybackslash}m{2cm}|>{\centering\arraybackslash}m{2cm}|>{\centering\arraybackslash}m{2cm}|}
\hline
\textbf{$L^2$ relative error} & \textbf{Variables} & \textbf{FME-PINNs} & \textbf{FME-KANs} \\ \hline
\multirow{4}{*}{} & $C_a$ & 6.181\% & 7.950\% \\
 & $C_b$ & 1.920\% & 3.163\% \\
 & T & 2.182\% & 1.731\% \\ 
& $E_a$ & 1.198\% & 2.112\% \\ \hline
\end{tabular}
\caption{Comparison of the $L^2$ relative errors (in percent) of FME-PINNs and FME-KANs for each variable ($C_a, C_b, T$) and unknown kinetic parameter ($E_a$).}
\label{tab: inverse_kan}
\end{table}

\subsection{Ablation Study}
To show the effectiveness of the FMEnets framework in solving reactive flow problems, we conduct an ablation study to evaluate the contributions of various enhancements in the FMEnets framework. This ablation study was conducted by systematically removing key components, including weight normalization, residual-based attention weights (RBA), and sequential training for the inverse problem. Table~\ref{tab:ablation_study} presents the relative \(L^2\) error for species concentrations (\(C_a, C_b, C_c\)) and temperature (\(T\)) across different model configurations for three-component reaction systems. The FME-PINNs configuration achieves the lowest error across all variables, demonstrating the full architecture's effectiveness. Removing weight normalization or output transformation leads to a slight increase in error. The output transformation here refers to the constrained domain for the concentration and temperature. This means that even though we don't know the range of concentration and temperature, FMEnets can also give us accurate predictions. We found that the RBA as a local weight strategy is significant in optimizing these stiff and complex multi-physics problems. However, we believe other comparable local weighting or sampling methods to RBA will also provide the support we need for the FMEnets. The most significant error increase occurs when the FMEnets architecture is entirely removed. At this stage, it becomes PINNs with enhancements, including weight normalization, output transformation, and RBA. This proves that the proposed FMEnets structure is the key to successfully modeling this type of reactive flow problem. In the inverse problem setting, excluding sequential training in the inverse model significantly increases errors. Our results indicate that both iteration-based global weighting and sequential training are essential for reliable kinetic parameter estimation and concentration prediction in reactive flow systems.

\begin{table}[h]
    \centering
    \begin{tabular}{lcccc}
        \hline
        \textbf{Mode} & \(\mathbf{C_a}\) & \(\mathbf{C_b}\) & \(\mathbf{C_c}\) & \(\mathbf{T}\)\\
        \hline
        FME-PINNs & 2.334\% & 2.734\% & 0.991\% & 1.206\%\\
        w/o Weight Normalization & 2.280\% & 3.232\% & 1.248\% & 1.230\% \\
        w/o Output Transformation & 3.595\% & 3.790\% & 2.247\% & 0.982\% \\
        w/o RBA & 13.190\% & 19.920\% & 9.395\% & 3.746\% \\
        w/o FMEnets architecture & 177.300\% & 96.610\% & 54.880\% & 8.035\% \\
        \hline\hline
        Inverse FME-PINNs & 7.920\% & 8.407\% & 3.647\% & 1.685\% \\
        (Inverse) w/o sequential training & 21.790\% & 11.490\% & 3.035\% & 3.238\% \\
        \hline
    \end{tabular}
    \caption{Relative \(L^2\) error for each case in the single-component ablation study of the three-component reaction system.}
    \label{tab:ablation_study}
\end{table}

\section{Summary}
In this study, we introduced FMEnets, a new physics-informed machine learning framework that seamlessly integrates the Navier–Stokes, material balance, and energy balance equations. FMEnets is composed of three interconnected sub-networks, each responsible for enforcing one of these governing equations. A sequential training strategy is used to facilitate the transfer of coupling variables among sub-networks. Each sub-network can be implemented using either a multilayer perceptron (MLP) or a Kolmogorov–Arnold Network (KAN), resulting in two variants: FME-PINNs and FME-KANs.

FMEnets addresses both forward and inverse problems in reactor design. In the forward mode, we applied FME-PINNs to three reactor scenarios involving highly exothermic reactions in non-ideal plug flow reactors immersed in an isothermal bath. Using only inlet and outlet data, FME-PINNs accurately predict velocity and pressure (less than 1.0\% relative error), species concentrations (0.991\% - 7.952\% relative error), and temperature profiles (less than 1.7\% relative error). Furthermore, we conducted a comparative study of FME-KANs and FME-PINNs at varying noise levels (1\% to 10\%) at different reactor locations (inlet, outlet and both). Even at 10\% noise in both inlet and outlet data, FME-KANs consistently demonstrated robust performance, maintaining less than 6\% relative error across all state variables. In the inverse mode, we assumed that the activation energy of the reactions are unknown and incorporated sparse, multi-residence-time experimental observations to infer the kinetic parameters. FMEnets achieved high accuracy for this inverse problem, with relative errors under 2.5\%. Comprehensive ablation studies further confirmed that the FMEnets architecture is critical for achieving robust, high-accuracy predictions, while residual-based attention weighting is essential for enhancing accuracy in this multiphysics setting.

Overall, FMEnets is the first physics-informed machine learning model that bridges flow, reactive transport, and energy transfer equations with empirical correlations and sparse, noisy data. This integration significantly expands the flexibility and capabilities of machine learning for rapid reactor design and optimization in complex chemical processes. Future work will focus on extending the framework to accommodate experimental equipment constraints and to incorporate more realistic reactor configurations and reaction networks, thereby further broadening its industrial applicability.

\section{Acknowledgments}
This work was supported by a Takeda gift to Brown University (BU). Partial support to the BU co-authors was also provided by the projects: NIH R01AT012312, R01HL168473,  and ONR Vannevar Bush N00014-22-1-2795.

\bibliographystyle{unsrt}
\bibliography{main}

\begin{thebibliography}{10}

\bibitem{fuad2015systematic}
Mohd Nazri~Mohd Fuad and Mohd~Azlan Hussain.
\newblock Systematic design of chemical reactors with multiple stages via multi-objective optimization approach.
\newblock In {\em Computer Aided Chemical Engineering}, volume~37, pages 869--874. Elsevier, 2015.

\bibitem{sheldon2012fundamentals}
Roger~A Sheldon.
\newblock Fundamentals of green chemistry: efficiency in reaction design.
\newblock {\em Chemical Society Reviews}, 41(4):1437--1451, 2012.

\bibitem{froment1979chemical}
Gilbert~F Froment.
\newblock {\em Chemical reactor analysis and design}.
\newblock 1979.

\bibitem{towler2021chemical}
Gavin Towler and Ray Sinnott.
\newblock {\em Chemical engineering design: principles, practice and economics of plant and process design}.
\newblock Butterworth-Heinemann, 2021.

\bibitem{stoessel2021thermal}
Francis Stoessel.
\newblock {\em Thermal safety of chemical processes: risk assessment and process design}.
\newblock John Wiley \& Sons, 2021.

\bibitem{fogler2010essentials}
H~Scott Fogler.
\newblock {\em Essentials of chemical reaction engineering: essenti chemica reactio engi}.
\newblock Pearson education, 2010.

\bibitem{zhu2022review}
Li-Tao Zhu, Xi-Zhong Chen, Bo~Ouyang, Wei-Cheng Yan, He~Lei, Zhe Chen, and Zheng-Hong Luo.
\newblock Review of machine learning for hydrodynamics, transport, and reactions in multiphase flows and reactors.
\newblock {\em Industrial \& Engineering Chemistry Research}, 61(28):9901--9949, 2022.

\bibitem{haghighatlari2019advances}
Mojtaba Haghighatlari and Johannes Hachmann.
\newblock Advances of machine learning in molecular modeling and simulation.
\newblock {\em Current Opinion in Chemical Engineering}, 23:51--57, 2019.

\bibitem{wu2023application}
Zhiyong Wu, Huan Wang, Chang He, Bingjian Zhang, Tao Xu, and Qinglin Chen.
\newblock The application of physics-informed machine learning in multiphysics modeling in chemical engineering.
\newblock {\em Industrial \& Engineering Chemistry Research}, 62(44):18178--18204, 2023.

\bibitem{ghasem2024combining}
Nayef Ghasem.
\newblock Combining cfd and ai/ml modeling to improve the performance of polypropylene fluidized bed reactors.
\newblock {\em Fluids}, 9(12):298, 2024.

\bibitem{li2023analysis}
Guansheng Li, Ting Ye, Zehong Xia, Sitong Wang, and Ziwei Zhu.
\newblock Analysis and prediction of hematocrit in microvascular networks.
\newblock {\em International Journal of Engineering Science}, 191:103901, 2023.

\bibitem{cao2024laplace}
Qianying Cao, Somdatta Goswami, and George~Em Karniadakis.
\newblock Laplace neural operator for solving differential equations.
\newblock {\em Nature Machine Intelligence}, 6(6):631--640, 2024.

\bibitem{wang2021learning}
Sifan Wang, Hanwen Wang, and Paris Perdikaris.
\newblock Learning the solution operator of parametric partial differential equations with physics-informed deeponets.
\newblock {\em Science advances}, 7(40):eabi8605, 2021.

\bibitem{karniadakis2021physics}
George~Em Karniadakis, Ioannis~G Kevrekidis, Lu~Lu, Paris Perdikaris, Sifan Wang, and Liu Yang.
\newblock Physics-informed machine learning.
\newblock {\em Nature Reviews Physics}, 3(6):422--440, 2021.

\bibitem{raissi2019physics}
Maziar Raissi, Paris Perdikaris, and George~E Karniadakis.
\newblock Physics-informed neural networks: A deep learning framework for solving forward and inverse problems involving nonlinear partial differential equations.
\newblock {\em Journal of Computational physics}, 378:686--707, 2019.

\bibitem{lu2021deepxde}
Lu~Lu, Xuhui Meng, Zhiping Mao, and George~Em Karniadakis.
\newblock Deepxde: A deep learning library for solving differential equations.
\newblock {\em SIAM review}, 63(1):208--228, 2021.

\bibitem{cuomo2022scientific}
Salvatore Cuomo, Vincenzo~Schiano Di~Cola, Fabio Giampaolo, Gianluigi Rozza, Maziar Raissi, and Francesco Piccialli.
\newblock Scientific machine learning through physics--informed neural networks: Where we are and what’s next.
\newblock {\em Journal of Scientific Computing}, 92(3):88, 2022.

\bibitem{toscano2025pinns}
Juan~Diego Toscano, Vivek Oommen, Alan~John Varghese, Zongren Zou, Nazanin Ahmadi~Daryakenari, Chenxi Wu, and George~Em Karniadakis.
\newblock From pinns to pikans: Recent advances in physics-informed machine learning.
\newblock {\em Machine Learning for Computational Science and Engineering}, 1(1):1--43, 2025.

\bibitem{hao2022physics}
Zhongkai Hao, Songming Liu, Yichi Zhang, Chengyang Ying, Yao Feng, Hang Su, and Jun Zhu.
\newblock Physics-informed machine learning: A survey on problems, methods and applications.
\newblock {\em arXiv preprint arXiv:2211.08064}, 2022.

\bibitem{chen2020physics}
Yuyao Chen, Lu~Lu, George~Em Karniadakis, and Luca Dal~Negro.
\newblock Physics-informed neural networks for inverse problems in nano-optics and metamaterials.
\newblock {\em Optics express}, 28(8):11618--11633, 2020.

\bibitem{niu2023modeling}
Sijun Niu, Enrui Zhang, Yuri Bazilevs, and Vikas Srivastava.
\newblock Modeling finite-strain plasticity using physics-informed neural network and assessment of the network performance.
\newblock {\em Journal of the Mechanics and Physics of Solids}, 172:105177, 2023.

\bibitem{raissi2020hidden}
Maziar Raissi, Alireza Yazdani, and George~Em Karniadakis.
\newblock Hidden fluid mechanics: Learning velocity and pressure fields from flow visualizations.
\newblock {\em Science}, 367(6481):1026--1030, 2020.

\bibitem{cai2021physics}
Shengze Cai, Zhiping Mao, Zhicheng Wang, Minglang Yin, and George~Em Karniadakis.
\newblock Physics-informed neural networks (pinns) for fluid mechanics: A review.
\newblock {\em Acta Mechanica Sinica}, 37(12):1727--1738, 2021.

\bibitem{zhang2023artificial}
Qian Zhang, Chenxi Wu, Adar Kahana, Youngeun Kim, Yuhang Li, George~Em Karniadakis, and Priyadarshini Panda.
\newblock Artificial to spiking neural networks conversion for scientific machine learning.
\newblock {\em arXiv preprint arXiv:2308.16372}, 2023.

\bibitem{sharma2023review}
Pushan Sharma, Wai~Tong Chung, Bassem Akoush, and Matthias Ihme.
\newblock A review of physics-informed machine learning in fluid mechanics.
\newblock {\em Energies}, 16(5):2343, 2023.

\bibitem{yazdani2020systems}
Alireza Yazdani, Lu~Lu, Maziar Raissi, and George~Em Karniadakis.
\newblock Systems biology informed deep learning for inferring parameters and hidden dynamics.
\newblock {\em PLoS computational biology}, 16(11):e1007575, 2020.

\bibitem{daneker2023systems}
Mitchell Daneker, Zhen Zhang, George~Em Karniadakis, and Lu~Lu.
\newblock Systems biology: Identifiability analysis and parameter identification via systems-biology-informed neural networks.
\newblock In {\em Computational Modeling of Signaling Networks}, pages 87--105. Springer, 2023.

\bibitem{zapf2022investigating}
Bastian Zapf, Johannes Haubner, Miroslav Kuchta, Geir Ringstad, Per~Kristian Eide, and Kent-Andre Mardal.
\newblock Investigating molecular transport in the human brain from mri with physics-informed neural networks.
\newblock {\em Scientific Reports}, 12(1):15475, 2022.

\bibitem{lu2021physics}
Lu~Lu, Raphael Pestourie, Wenjie Yao, Zhicheng Wang, Francesc Verdugo, and Steven~G Johnson.
\newblock Physics-informed neural networks with hard constraints for inverse design.
\newblock {\em SIAM Journal on Scientific Computing}, 43(6):B1105--B1132, 2021.

\bibitem{wang2024digital}
Danshi Wang, Yuchen Song, Yao Zhang, Xiaotian Jiang, Jiawei Dong, Faisal~Nadeem Khan, Takeo Sasai, Shanguo Huang, Alan Pak~Tao Lau, Massimo Tornatore, et~al.
\newblock Digital twin of optical networks: a review of recent advances and future trends.
\newblock {\em Journal of Lightwave Technology}, 2024.

\bibitem{yang2024data}
Sunwoong Yang, Hojin Kim, Yoonpyo Hong, Kwanjung Yee, Romit Maulik, and Namwoo Kang.
\newblock Data-driven physics-informed neural networks: A digital twin perspective.
\newblock {\em Computer Methods in Applied Mechanics and Engineering}, 428:117075, 2024.

\bibitem{ding2023self}
Yi~Ding, Su~Chen, Xiaojun Li, Suyang Wang, Shaokai Luan, and Hao Sun.
\newblock Self-adaptive physics-driven deep learning for seismic wave modeling in complex topography.
\newblock {\em Engineering Applications of Artificial Intelligence}, 123:106425, 2023.

\bibitem{patel2023optimal}
Rahul Patel, Sharad Bhartiya, and Ravindra Gudi.
\newblock Optimal temperature trajectory for tubular reactor using physics informed neural networks.
\newblock {\em Journal of Process Control}, 128:103003, 2023.

\bibitem{cohen2024data}
Ben Cohen, Burcu Beykal, and George~M Bollas.
\newblock Data-driven discovery of reaction kinetic models in dynamic plug flow reactors using symbolic regression.
\newblock In {\em Computer Aided Chemical Engineering}, volume~53, pages 2947--2952. Elsevier, 2024.

\bibitem{choi2022physics}
Solji Choi, Ikhwan Jung, Haeun Kim, Jonggeol Na, and Jong~Min Lee.
\newblock Physics-informed deep learning for data-driven solutions of computational fluid dynamics.
\newblock {\em Korean Journal of Chemical Engineering}, 39(3):515--528, 2022.

\bibitem{li2024unit}
Haochen Li, David Spelman, and John Sansalone.
\newblock Unit operation and process modeling with physics-informed machine learning.
\newblock {\em Journal of Environmental Engineering}, 150(4):04024002, 2024.

\bibitem{qiu2022physics}
Rundi Qiu, Renfang Huang, Yao Xiao, Jingzhu Wang, Zhen Zhang, Jieshun Yue, Zhong Zeng, and Yiwei Wang.
\newblock Physics-informed neural networks for phase-field method in two-phase flow.
\newblock {\em Physics of Fluids}, 34(5), 2022.

\bibitem{lastrucci2024physics}
Giacomo Lastrucci, Maximilian~F Theisen, and Artur~M Schweidtmann.
\newblock Physics-informed neural networks and time-series transformer for modeling of chemical reactors.
\newblock In {\em Computer aided chemical engineering}, volume~53, pages 571--576. Elsevier, 2024.

\bibitem{wu2024physics}
Zhiyong Wu, Mingjian Li, Chang He, Bingjian Zhang, Jingzheng Ren, Haoshui Yu, and Qinglin Chen.
\newblock Physics-informed learning of chemical reactor systems using decoupling--coupling training framework.
\newblock {\em AIChE Journal}, 70(7):e18436, 2024.

\bibitem{hou2025physics}
Qingzhi Hou, Xiaolong Xu, Zewei Sun, Jianping Wang, and Vijay~P Singh.
\newblock Physics informed neural network for forward and inverse multispecies contaminant transport with variable parameters.
\newblock {\em Journal of Hydrology}, 655:132977, 2025.

\bibitem{sun2023physics}
Zewei Sun, Honghan Du, Chunfu Miao, and Qingzhi Hou.
\newblock A physics-informed neural network based simulation tool for reacting flow with multicomponent reactants.
\newblock {\em Advances in Engineering Software}, 185:103525, 2023.

\bibitem{peng2023physics}
Jiang-Zhou Peng, Yue Hua, Yu-Bai Li, Zhi-Hua Chen, Wei-Tao Wu, and Nadine Aubry.
\newblock Physics-informed graph convolutional neural network for modeling fluid flow and heat convection.
\newblock {\em Physics of Fluids}, 35(8), 2023.

\bibitem{laubscher2021simulation}
Ryno Laubscher.
\newblock Simulation of multi-species flow and heat transfer using physics-informed neural networks.
\newblock {\em Physics of Fluids}, 33(8), 2021.

\bibitem{rajendra2022advancement}
P~Rajendra, A~Girisha, and T~Gunavardhana Naidu.
\newblock Advancement of machine learning in materials science.
\newblock {\em Materials Today: Proceedings}, 62:5503--5507, 2022.

\bibitem{brunton2020machine}
Steven~L Brunton, Bernd~R Noack, and Petros Koumoutsakos.
\newblock Machine learning for fluid mechanics.
\newblock {\em Annual review of fluid mechanics}, 52(1):477--508, 2020.

\bibitem{wei2019machine}
Jing Wei, Xuan Chu, Xiang-Yu Sun, Kun Xu, Hui-Xiong Deng, Jigen Chen, Zhongming Wei, and Ming Lei.
\newblock Machine learning in materials science.
\newblock {\em InfoMat}, 1(3):338--358, 2019.

\bibitem{sidey2019machine}
Jenni~AM Sidey-Gibbons and Chris~J Sidey-Gibbons.
\newblock Machine learning in medicine: a practical introduction.
\newblock {\em BMC medical research methodology}, 19:1--18, 2019.

\bibitem{wu2023gpt}
Chenxi Wu, Alan~John Varghese, Vivek Oommen, and George~Em Karniadakis.
\newblock Gpt vs human for scientific reviews: A dual source review on applications of chatgpt in science.
\newblock {\em Journal of Machine Learning for Modeling and Computing}, 2023.

\bibitem{ngo2022forward}
Son~Ich Ngo and Young-Il Lim.
\newblock Forward physics-informed neural networks suitable for multiple operating conditions of catalytic co2 methanation isothermal fixed-bed.
\newblock {\em IFAC-PapersOnLine}, 55(7):429--434, 2022.

\bibitem{arrhenius1889reaktionsgeschwindigkeit}
Svante Arrhenius.
\newblock {\"U}ber die reaktionsgeschwindigkeit bei der inversion von rohrzucker durch s{\"a}uren.
\newblock {\em Zeitschrift f{\"u}r physikalische Chemie}, 4(1):226--248, 1889.

\bibitem{shukla2024comprehensive}
Khemraj Shukla, Juan~Diego Toscano, Zhicheng Wang, Zongren Zou, and George~Em Karniadakis.
\newblock A comprehensive and fair comparison between mlp and kan representations for differential equations and operator networks.
\newblock {\em arXiv preprint arXiv:2406.02917}, 2024.

\bibitem{hornik1989multilayer}
Kurt Hornik, Maxwell Stinchcombe, and Halbert White.
\newblock {Multilayer feedforward networks are universal approximators}.
\newblock {\em Neural Networks}, 2(5):359--366, 1989.

\bibitem{liu2024kan}
Ziming Liu, Yixuan Wang, Sachin Vaidya, Fabian Ruehle, James Halverson, Marin Solja{\v{c}}i{\'c}, Thomas~Y Hou, and Max Tegmark.
\newblock Kan: Kolmogorov-arnold networks.
\newblock {\em arXiv preprint arXiv:2404.19756}, 2024.

\bibitem{toscano2024kkans}
Juan~Diego Toscano, Li-Lian Wang, and George~Em Karniadakis.
\newblock Kkans: Kurkova-kolmogorov-arnold networks and their learning dynamics.
\newblock {\em arXiv preprint arXiv:2412.16738}, 2024.

\bibitem{salimans2016weight}
Tim Salimans and Durk~P Kingma.
\newblock Weight normalization: A simple reparameterization to accelerate training of deep neural networks.
\newblock {\em Advances in neural information processing systems}, 29, 2016.

\bibitem{anagnostopoulos2024residual}
Sokratis~J Anagnostopoulos, Juan~Diego Toscano, Nikolaos Stergiopulos, and George~Em Karniadakis.
\newblock Residual-based attention in physics-informed neural networks.
\newblock {\em Computer Methods in Applied Mechanics and Engineering}, 421:116805, 2024.

\bibitem{dong2021method}
Suchuan Dong and Naxian Ni.
\newblock A method for representing periodic functions and enforcing exactly periodic boundary conditions with deep neural networks.
\newblock {\em Journal of Computational Physics}, 435:110242, 2021.

\bibitem{wang2021understanding}
Sifan Wang, Yujun Teng, and Paris Perdikaris.
\newblock Understanding and mitigating gradient flow pathologies in physics-informed neural networks.
\newblock {\em SIAM Journal on Scientific Computing}, 43(5):A3055--A3081, 2021.

\bibitem{chen2020comparison}
Jingrun Chen, Rui Du, and Keke Wu.
\newblock A comparison study of deep galerkin method and deep ritz method for elliptic problems with different boundary conditions.
\newblock {\em arXiv preprint arXiv:2005.04554}, 2020.

\bibitem{sukumar2022exact}
N~Sukumar and Ankit Srivastava.
\newblock Exact imposition of boundary conditions with distance functions in physics-informed deep neural networks.
\newblock {\em Computer Methods in Applied Mechanics and Engineering}, 389:114333, 2022.

\bibitem{mao2020physics}
Zhiping Mao, Ameya~D Jagtap, and George~Em Karniadakis.
\newblock {Physics-informed neural networks for high-speed flows}.
\newblock {\em Computer Methods in Applied Mechanics and Engineering}, 360:112789, 2020.

\bibitem{wu2023comprehensive}
Chenxi Wu, Min Zhu, Qinyang Tan, Yadhu Kartha, and Lu~Lu.
\newblock A comprehensive study of non-adaptive and residual-based adaptive sampling for physics-informed neural networks.
\newblock {\em Computer Methods in Applied Mechanics and Engineering}, 403:115671, 2023.

\bibitem{mcclenny2023self}
Levi~D McClenny and Ulisses~M Braga-Neto.
\newblock Self-adaptive physics-informed neural networks.
\newblock {\em Journal of Computational Physics}, 474:111722, 2023.

\bibitem{anagnostopoulos2024learning}
Sokratis~J Anagnostopoulos, Juan~Diego Toscano, Nikolaos Stergiopulos, and George~Em Karniadakis.
\newblock Learning in {PINNs}: Phase transition, total diffusion, and generalization.
\newblock {\em arXiv preprint arXiv:2403.18494}, 2024.

\bibitem{xiang2022self}
Zixue Xiang, Wei Peng, Xu~Liu, and Wen Yao.
\newblock Self-adaptive loss balanced physics-informed neural networks.
\newblock {\em Neurocomputing}, 496:11--34, 2022.

\bibitem{liu2021dual}
Dehao Liu and Yan Wang.
\newblock A dual-dimer method for training physics-constrained neural networks with minimax architecture.
\newblock {\em Neural Networks}, 136:112--125, 2021.

\bibitem{basir2022investigating}
Shamsulhaq Basir.
\newblock Investigating and mitigating failure modes in physics-informed neural networks (pinns).
\newblock {\em arXiv preprint arXiv:2209.09988}, 2022.

\bibitem{wang2022respecting}
Sifan Wang, Shyam Sankaran, and Paris Perdikaris.
\newblock Respecting causality is all you need for training physics-informed neural networks.
\newblock {\em arXiv preprint arXiv:2203.07404}, 2022.

\bibitem{wang2022and}
Sifan Wang, Xinling Yu, and Paris Perdikaris.
\newblock {When and why {PINNs}} fail to train: A neural tangent kernel perspective.
\newblock {\em Journal of Computational Physics}, 449:110768, 2022.

\bibitem{krishnapriyan2021characterizing}
Aditi Krishnapriyan, Amir Gholami, Shandian Zhe, Robert Kirby, and Michael~W Mahoney.
\newblock {{Characterizing possible failure modes in physics-informed neural networks}}.
\newblock {\em Advances in Neural Information Processing Systems}, 34:26548--26560, 2021.

\bibitem{zhang2024meshless}
Han Zhang, Raymond~H Chan, and Xue-Cheng Tai.
\newblock {A Meshless Solver for Blood Flow Simulations in Elastic Vessels Using a Physics-Informed Neural Network}.
\newblock {\em SIAM Journal on Scientific Computing}, 46(4):C479--C507, 2024.

\bibitem{wang2023expert}
Sifan Wang, Shyam Sankaran, Hanwen Wang, and Paris Perdikaris.
\newblock {An expert's guide to training physics-informed neural networks}.
\newblock {\em arXiv preprint arXiv:2308.08468}, 2023.

\bibitem{wang2024piratenets}
Sifan Wang, Bowen Li, Yuhan Chen, and Paris Perdikaris.
\newblock {PirateNets: Physics-informed Deep Learning with Residual Adaptive Networks}.
\newblock {\em arXiv preprint arXiv:2402.00326}, 2024.

\bibitem{ahmadi2024ai}
Nazanin Ahmadi~Daryakenari, Mario De~Florio, Khemraj Shukla, and George~Em Karniadakis.
\newblock {AI-Aristotle: A physics-informed framework for systems biology gray-box identification}.
\newblock {\em PLOS Computational Biology}, 20(3):e1011916, 2024.

\bibitem{daryakenari2025cminns}
Nazanin~Ahmadi Daryakenari, Shupeng Wang, and George Karniadakis.
\newblock Cminns: Compartment model informed neural networks—unlocking drug dynamics.
\newblock {\em Computers in Biology and Medicine}, 184:109392, 2025.

\bibitem{toscano2024invivo}
Juan~Diego Toscano, Chenxi Wu, Antonio Ladron-de Guevara, Ting Du, Maiken Nedergaard, Douglas~H Kelley, George~Em Karniadakis, and Kimberly Boster.
\newblock {Inferring in vivo murine cerebrospinal fluid flow using artificial intelligence velocimetry with moving boundaries and uncertainty quantification}.
\newblock {\em bioRxiv}, pages 2024--08, 2024.

\bibitem{toscano2024inferring}
Juan~Diego Toscano, Theo K{\"a}ufer, Martin Maxey, Christian Cierpka, and George~Em Karniadakis.
\newblock {Inferring turbulent velocity and temperature fields and their statistics from Lagrangian velocity measurements using physics-informed Kolmogorov-Arnold Networks}.
\newblock {\em arXiv preprint arXiv:2407.15727}, 2024.

\bibitem{rathore2024challenges}
Pratik Rathore, Weimu Lei, Zachary Frangella, Lu~Lu, and Madeleine Udell.
\newblock Challenges in training pinns: A loss landscape perspective.
\newblock {\em arXiv preprint arXiv:2402.01868}, 2024.

\bibitem{alnaes2015fem}
M.~S. Aln{\ae}s, J.~Blechta, J.~Hake, H.~A. Johansen, and K.~H. Karlsen.
\newblock The fenics project version 1.5.
\newblock {\em Archives of Computational Methods in Engineering}, 22(3):253--290, 2015.

\bibitem{petsc-user-ref}
Barry~F. Smith, Hong Zhang, et~al.
\newblock Petsc: Portable, extensible toolkit for scientific computation.
\newblock {\em ACM Transactions on Mathematical Software}, 32(3):669--702, 2006.

\bibitem{amestoy2001mumps}
P.~R. Amestoy, I.~S. Duff, and J.~Y. L'Excellent.
\newblock Mumps: A general purpose distributed memory sparse direct solver.
\newblock {\em Parallel Computing}, 30(2):235--274, 2004.

\end{thebibliography}

\clearpage 
\appendix
\section{Appendix} \label{Appendix}
\subsection{Physical and Chemical Properties of the Fluid}
\begin{table}[h!]
\centering
\begin{tabular}{lll}
\toprule
\textbf{Parameter} & \textbf{Value} & \textbf{Units} \\
\midrule
$\mu$   & 0.000654416 & N$\cdot$s/(m$^2$) \\
$C_p$   & 4200        & J/(kg$\cdot$K)    \\
$\rho$  & 993         & kg/m$^3$         \\
$D_{AB}$ & $1.038 \times 10^{-7}$ & m$^2$/s  \\
$K_c$   & 0.5         & W/(m$\cdot$K)    \\
\bottomrule
\end{tabular}
\end{table}

\subsection{Actual Dimensions of the Plug Flow Reactor}
\begin{table}[h!]
\centering
\begin{tabular}{lll}
\toprule
\textbf{Parameter} & \textbf{Value} & \textbf{Units} \\
\midrule
$r$   & 0.000793 & m \\
$z$   & 0.020612 & m   \\
\bottomrule
\end{tabular}
\end{table}

\subsection{Dimensionless Numbers}
\begin{table}[h!]
\centering
\begin{tabular}{ll}
\toprule
\textbf{Dimensionless Group} & \textbf{Value} \\
\midrule
Re   & 40.343   \\
Pe   & 256.008  \\
Pe$_T$ & 221.764 \\
\bottomrule
\end{tabular}
\end{table}

\subsection{Dimensionless Geometry}
\begin{table}[h!]
\centering
\begin{tabular}{ll}
\toprule
\textbf{Parameter} & \textbf{Value} \\
\midrule
Length & 12.984 \\
Radius & 0.5    \\
\bottomrule
\end{tabular}
\end{table}

\newpage
\subsection{Characteristic Dimensions}
\begin{table}[h!]
\centering
\begin{tabular}{ll}
\toprule
\textbf{Parameter} & \textbf{Value} \\
\midrule
$V$              & 0.0167  \\
$D$              & 0.00158 \\
$P_{\mathrm{char}}$         & 0.27851 \\
$T_{\mathrm{char}}$ & 300      \\
$C_{\mathrm{char}}$ & 1300     \\
\bottomrule
\end{tabular}
\end{table}

\subsection{Reactions, Kinetics and Initial Conditions}
\subsubsection{Case 1}

\begin{table}[h!]
\centering
\begin{tabular}{lccc}
\toprule
\textbf{Reaction} & \boldmath{$k_0$} & \boldmath{$E_a$} & \boldmath{$\Delta H$} \\
\midrule
$r_{A}$ & 400 m$^3$/(s$\cdot$mol) & 40230 J/mol & -100000 J/mol \\
\bottomrule
\end{tabular}
\end{table}

\begin{table}[h!]
\centering
\begin{tabular}{lccc}
\toprule
\textbf{Initial Conditions} & \boldmath{$C_a$} & \boldmath{$C_b$} & \boldmath{$T$} \\
\midrule
& 1300 mol/m$^3$ & 0 mol/m$^3$ & 400 K \\
\bottomrule
\end{tabular}
\end{table}

\subsubsection{Case 2}
\begin{table}[h!]
\centering
\begin{tabular}{lccc}
\toprule
\textbf{Reaction} & \boldmath{$k_0$} & \boldmath{$E_a$} & \boldmath{$\Delta H$} \\
\midrule
$r_{A}$ & 400 m$^3$/(s$\cdot$mol) & 40230 J/mol & -100000 J/mol \\
$r_{B}$ & 400 m$^3$/(s$\cdot$mol) & 40230 J/mol & -100000 J/mol \\
\bottomrule
\end{tabular}
\end{table}

\begin{table}[h!]
\centering
\begin{tabular}{lcccc}
\toprule
\textbf{Initial Conditions} & \boldmath{$C_a$} & \boldmath{$C_b$} & \boldmath{$C_c$} & \boldmath{$T$} \\
\midrule
& 1300 mol/m$^3$ & 0 mol/m$^3$ & 0 mol/m$^3$ & 400 K \\
\bottomrule
\end{tabular}
\end{table}

\newpage
\subsubsection{Case 3}
\begin{table}[h!]
\centering
\begin{tabular}{lccc}
\toprule
\textbf{Reaction} & \boldmath{$k_0$} & \boldmath{$E_a$} & \boldmath{$\Delta H$} \\
\midrule
$r_A$ & 284 m$^3$/(s$\cdot$mol) & 40230 J/mol & -100000 J/mol \\
$r_B$ & 142 m$^3$/(s$\cdot$mol) & 40230 J/mol & -100000 J/mol \\
$r_C$ & 227 m$^3$/(s$\cdot$mol) & 40230 J/mol & -100000 J/mol \\
\bottomrule
\end{tabular}
\end{table}

\begin{table}[h!]
\centering
\small
\begin{tabular}{lccccc}
\toprule
\textbf{Initial Conditions} & \boldmath{$C_a$} & \boldmath{$C_b$} & \boldmath{$C_c$} & \boldmath{$C_d$-$C_f$} & \boldmath{$T$} \\
\midrule
& 850 mol/m$^3$ & 1300 mol/m$^3$ & 1250 mol/m$^3$ & 0 mol/m$^3$ & 400 K \\
\bottomrule
\end{tabular}
\end{table}

\end{document}